\pdfoutput = 1

\documentclass{article}

\usepackage{microtype}
\usepackage{graphicx}
\usepackage{subfigure}
\usepackage{booktabs} % for professional tables
\usepackage{amsmath,amssymb}
\usepackage{url}
\DeclareMathOperator*{\argmin}{arg\,min}
\usepackage{multirow}
\usepackage{hyperref}
\usepackage{color}

\usepackage[accepted]{icml2019}

\icmltitlerunning{QGAN: Quantized Generative Adversarial Networks}

\begin{document}

\twocolumn[
\icmltitle{QGAN: Quantized Generative Adversarial Networks}





%\author{
%\begin{tabular}{c}
%Peiqi Wang$^1$, Dongsheng Wang$^1$, Yu Ji$^1$, Xinfeng Xie$^2$, Haoxuan Song$^1$, Xuxin Liu$^1$, Yongqiang Lyu$^1$, Yuan Xie$^2$\\
%\vspace{1pt}
%\end{tabular}\\
%\begin{tabular}{c}
%$^1$Department of Computer Science and Technology, Tsinghua University\\
%$^2$Department of Electrical and Computer Engineering, University of California, Santa Barbara
%\vspace{2pt}
%\end{tabular}\\
%\vspace{1pt}
%\begin{tabular}{c}
%\texttt{wds@mail.tsinghua.edu.cn}, ~\texttt{leideng@ucsb.edu}, ~\texttt{wpq14@mails.tsinghua.edu.cn}
%\texttt{wpq14@mails.tsinghua.edu.cn}, ~\texttt{wds@mail.tsinghua.edu.cn}\\
%\texttt{\{xinfeng, leideng, yuanxie\}@ucsb.edu}
%\end{tabular}
%}

\begin{icmlauthorlist}
\icmlauthor{Peiqi Wang}{thu}
\icmlauthor{Dongsheng Wang}{thu}
\icmlauthor{Yu Ji}{thu}
\icmlauthor{Xinfeng Xie}{ucsb}
\icmlauthor{Haoxuan Song}{thu} \\
\icmlauthor{Xuxin Liu}{thu}
\icmlauthor{Yongqiang Lyu}{thu}
\icmlauthor{Yuan Xie}{ucsb}
\end{icmlauthorlist}

{
    \vspace{2pt}
    \centering
    \begin{tabular}{c}
    $^1$Department of Computer Science and Technology, Tsinghua University\\
    $^2$Department of Electrical and Computer Engineering, University of California, Santa Barbara \\
    \texttt{wpq14@mails.tsinghua.edu.cn, wds@mail.tsinghua.edu.cn}
    \end{tabular} \\
}
\icmlaffiliation{thu}{Department of Computer Science and Technology, Tsinghua University, Beijing, China}
\icmlaffiliation{ucsb}{Department of Electrical and Computer Engineering, University of California, Santa Barbara, USA}

%\icmlcorrespondingauthor{Peiqi Wang}{wpq14@mails.tsinghua.edu.cn}
%\icmlcorrespondingauthor{Dongsheng Wang}{wds@mail.tsinghua.edu.cn}

\icmlkeywords{Machine Learning, ICML}

\vskip 0.3in
]




\begin{abstract}

The intensive computation and memory requirements of generative adversarial neural networks (GANs) hinder its real-world deployment on edge devices such as smartphones. Despite the success in model reduction of CNNs, neural network quantization methods have not yet been studied on GANs, which are mainly faced with the issues of both the effectiveness of quantization algorithms and the instability of training GAN models.
In this paper, we start with an extensive study on applying existing successful methods to quantize GANs.
Our observation reveals that none of them generates samples with reasonable quality because of the underrepresentation of quantized values in model weights, and the generator and discriminator networks show different sensitivities upon quantization methods. 
Motivated by these observations, we develop a novel quantization method for GANs based on EM algorithms, named as QGAN. 
We also propose a multi-precision algorithm to help find the optimal number of bits of quantized GAN models in conjunction with corresponding result qualities.
Experiments on CIFAR-10 and CelebA show that QGAN can quantize GANs to even 1-bit or 2-bit representations with results of quality comparable to original models.
\end{abstract}
%Motivated by these observations, we develop a novel quantization method, using an EM-based algorithm and a multi-precision quantization process to \peiqi{ support a trade-off between the number of bits and the quality of generated samples.}
% Experiments show \xinfeng{GAN models} quantized to 2-bit can generate samples with reasonable quality with the help of our novel quantization method and multi-precision quantization process.
\section{Introduction}

Generative adversarial networks (GANs) have obtained impressive success in a wide range of applications, such as super-resolution image generation, speech synthesis, image-to-image translation, and video frame prediction \cite{application1, application2}.
Despite their success in generating high-quality samples, these models are hard to be deployed into real-world applications on edge devices because of their huge demands for computation and storage capacity.
%\xinfeng{
For example, the BigGAN model~\cite{BigGAN}, developed by Google, contains up to 0.2 TOPs in the inference phase and its model size is over 1.3 GB.%}
This challenge becomes more urgent as the growth of privacy and security concerns about running the inference on cloud platforms.

%Generative models have evolved to the promising methods for a wide range of applications, like super-resolution images generation, speech synthesis, image to image translation, video frame prediction, and so on. 
%Generative adversarial network (GAN) \cite{GAN}is one of the prominent dominant approaches for generating samples and has achieved significant performance. 
%Some large GAN models\cite{PGGAN, BigGAN} show impressive results with great quality and also improve the stability of the models themselves. 
%Though these networks are very powerful, the huge demands of computation and storage limit their deployment. 
%For example, the BigGAN model \cite{BigGAN} proposed by Google is trained on 128 to 512 cores of a Google TPU, and the model size is over 1.3GB. 
%These not only make the large model is difficult to be reproduced by researchers, but also make it difficult to deploy GANs on the mobile system.

% Related work on Quantization and their shortcomings
% The challenges
% My contribution 

%\xinfeng{
This challenge exists in the deployment of various neural network models besides GANs.
State-of-the-art techniques to compress model scales include pruning, quantizaion, low-rank approximate ~\cite{deepcompression, bnn,lowrank}.
%, low-rank approximation~\cite{lowrank}, sparsity utilization~\cite{wenwei-nips}, and quantization~\cite{bnn}.
Among these techniques, quantization is the most easy-to-use and scalable method, which uses a less number of bits for data representations than a 32-bit single-precision.
Quantization has the following three aspects of advantages over other techniques.
First, the compression rate is significant.
For example, the model size of a 2-bit quantized model is reduced by 16$\times$.
Second, the quantization technique does not change neural network architectures. Thus it is orthogonal to the study of algorithms for neural network architecture exploration.
Finally, it is easy to be deployed into off-the-shelf devices with little hardware co-design to obtain significant performance and energy benefits.
The use of quantization methods requires little knowledge from algorithm researchers to hardware.

Although quantizing neural network models has achieved impressive success on convolutional neural networks (CNNs) and recurrent neural networks (RNNs)~\cite{dorefa-net, xnor-net}, there is still no successful attempt to quantize GAN models.
In this paper, we first study the effectiveness of typical quantization methods on GAN models.
Despite the success of these methods on CNNs or RNNs, We observe that they are not directly applicable to quantize GAN models because of the underrepresentation to original values.
%Besides, we observe that different components from GAN models have different sensitivities to the number of bits used in quantized values.
Besides, we observe features of the convergence and sensitivity of quantized GAN model.
%Besides, we observe that the discriminator is more sensitive than generator networks to the number of bits used in data representation. 
%During quantization process, a quantized discriminator can converge is a necessary and sufficient condition for the entir quantized GAN to converge.
Based on our observations, we develop QGAN, a novel quantization method based on Expectation{-}Maximization (EM) algorithm, and a novel multi-precision quantization algorithm.
Finally, our experiments show that the proposed Q can compress GAN models into 1-bit or 2-bit representations while generating samples of comparable quality, and our multi-precision method helps further improve the results quality of quantized GAN models according to a given demand.

In summary, our work has following contributions:

%\xinfeng{
\begin{itemize}
    \item We conduct an extensive study on the effectiveness of existing widely-used quantization methods. This empirical study demonstrates that these quantization methods are not applicable to GAN models directly although they work well on CNNs or RNNs.
    \item We obtain some observations of quantized GAN. First, the discriminator is more sensitive than the generator to the number of quantized bits. Second, a converged quantized discriminator can ensure the convergence of the entire quantized GAN model. Third, quantizing both the discriminator and generator is more stable than only quantizing generator networks.
    \item We propose QGAN, a novel quantization method for GAN models based on EM algorithm to overcome the data underrepresentation problem of existing quantization methods. Our experiments demonstrate that GAN models quantized to 2-bit or even 1-bit by QGAN can generate samples of comparable quality.
    \item We develop a multi-precision quantization algorithm based on our observations on the convergence and sensitivity of quantized GAN models. This algorithm provides the optimal number of bits configuration to satisfy the results quality requirement.
%    \item Our extensive experiments on various GAN models demonstrate that our quantization methods are generally applicable to a wide range of GAN models, including XXX (please fill in the name of GAN models here). 
\end{itemize}
%}

%\begin{itemize}
%    \item We provide comprehensive comparisons and analysis of the impact of different quantization methods on GANs. It empirically demonstrates the quantization methods which work well on CNNs are not applicable to GANs.
%    \item We propose a fine-grained quantization method based on EM algorithm, which can quantize GAN models to 2 bits with acceptable performance. It opens the way to deploy large GAN models to the mobile or embedded system.
%    \item On the basis of adversarial structure in GAN models, we introduce a multi-precision quantization method. We adopt different bitwidth on the generator and discriminator to further close the gap between the quantized model and the original one. 
%\end{itemize}

\section{Background}
\subsection{Generative adversarial networks}

%\xinfeng{
Generative Adversarial Network (GAN) is composed of two components, the generator and the discriminator.
The generator network, usually denoted as $G$ is trained to generate samples in a similar distribution of real data while the discriminator network, usually denoted as $D$ is trained to discriminate whether the input is generated by $G$ or from real data.
The generator takes a sampled noise $z$, where $z\sim\mathcal{N}(0,1) $ or $U(-1,1)$, as the input each time to generate a data sample.
Both samples generated from $G$ and real data are taken as inputs, denoted as $x$, to the discriminator, and the discriminator estimate the probability, $D(x)$, that the input is from real data.
The training process of a GAN model can be formulated as a min-max game between the generator and the discriminator.
The objective function of this min-max game can be formulated as:
\begin{equation}
\label{equ:GAN}
\begin{split}
    \min_G\max_D V(D,G) = &\mathbb{E}_{x\sim p_{data}(\textbf{x})}[\log D(\textbf{x})] + \\ &\mathbb{E}_{\textbf{z}\sim p(\textbf{z})}[\log (1-G(\textbf{z})]
\end{split}
\end{equation}

The generator aims to minimize this objective function while the discriminator aims to maximize it.
Both of them converge at a Nash equilibrium point where neither of them has any better action to further improve objects.

To improve the quality of generated samples, prior studies focus on better neural network architectures~\cite{DCGAN, CGAN, PGGAN}.
Some studies propose new objective functions for better convergence of the training process, such as adding new constraints~\cite{WGAN, improvedWGAN} and using smoother non-vanishing or non-exploding gradients~\cite{LSGAN, EBGAN}.
Our work focuses on using a smaller number of bits for data representations in GAN models for a more efficient deployment on edge devices, thus our work is orthogonal to these prior studies.
%}

\subsection{Quantizaion}

Quantization is a promising technique to reduce neural network model size and simplify arithmetic operations by reducing the number of bits in the data representation.
For example, in binary neural networks~\cite{bnn}, both weights and activations are quantized to 1-bit from original full-precision representation (32-bit).
In this case, the size of the model is reduced by 32$\times$ and the floating-point arithmetic operations are simplified into single-bit logical operations~\cite{xnor-net}.
From the perspective of hardware, these operations are easier to be implemented with higher performance, better energy efficiency, and smaller area overheads.
Therefore, these quantized models are easier to be deployed on edge devices because of smaller model sizes and hardware-friendly operations.

The benefits of quantization motivate prior research studies in CNNs and RNNs.
Among these studies, Binarized Neural Network \cite{bnn} uses a single sign function with a scaling factor to binarize the weights and activations. 
XNOR-Net \cite{xnor-net} formulates the quantization as an optimization problem, and successfully quantize CNNs to a binary neural network without accuracy loss. 
DoReFa-Net \cite{dorefa-net} adopts heuristic linear quantization to weights, activations, and gradients. 
%It works well on both CNNs and RNNs. 
Some studies pay attention to quantizing networks to extreme low bits \cite{bnn, TTQ, HitNet}, and some focus on quantizing more objects \cite{micikevicius2017mixed, banner2018scalable}, e.g. gradients, errors, and weight update. 
These studies have achieved great success by using an impressive low number of bits, usually 1-bit or 2-bit, while obtaining accuracy comparable to full-precision baseline models.

Despite these successful results on CNN and RNN, our work is the first to focus on quantizing GAN models.
Our case study in Section~\ref{sec:analysis} shows that these typical methods used in CNN and RNN models are not directly applicable to quantize GAN.
In this work, we develop a better quantization method based on the EM algorithm and a multi-precision training process for improving the quality of generated samples to meet specific quality demands.
%\peiqi{a more flexible trade-off} between the number of bits and the quality of generated samples.}

%\section{Analysis of the Quantization}
%\xinfeng{
\section{Study on Quantization Methods}
%}

\begin{figure*}[!t]
    \centering
    \subfigure[weights in D]{
        \label{fig:dcgan_d}
        \includegraphics[width=0.23\linewidth]{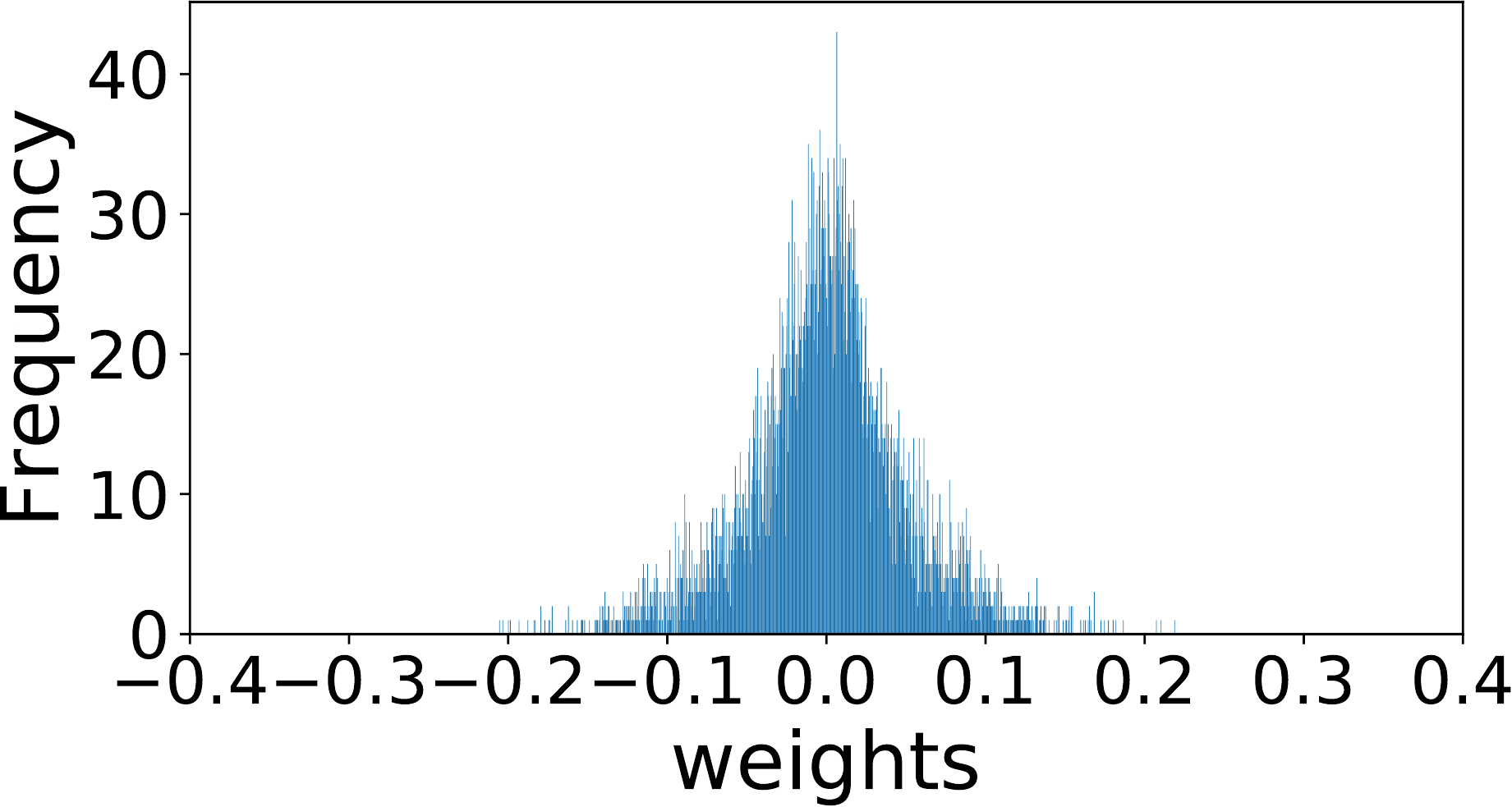} 
    }
    \subfigure[2-bit minmax-Q in D]{
        \label{fig:minmax_d}
        \includegraphics[width=0.23\linewidth]{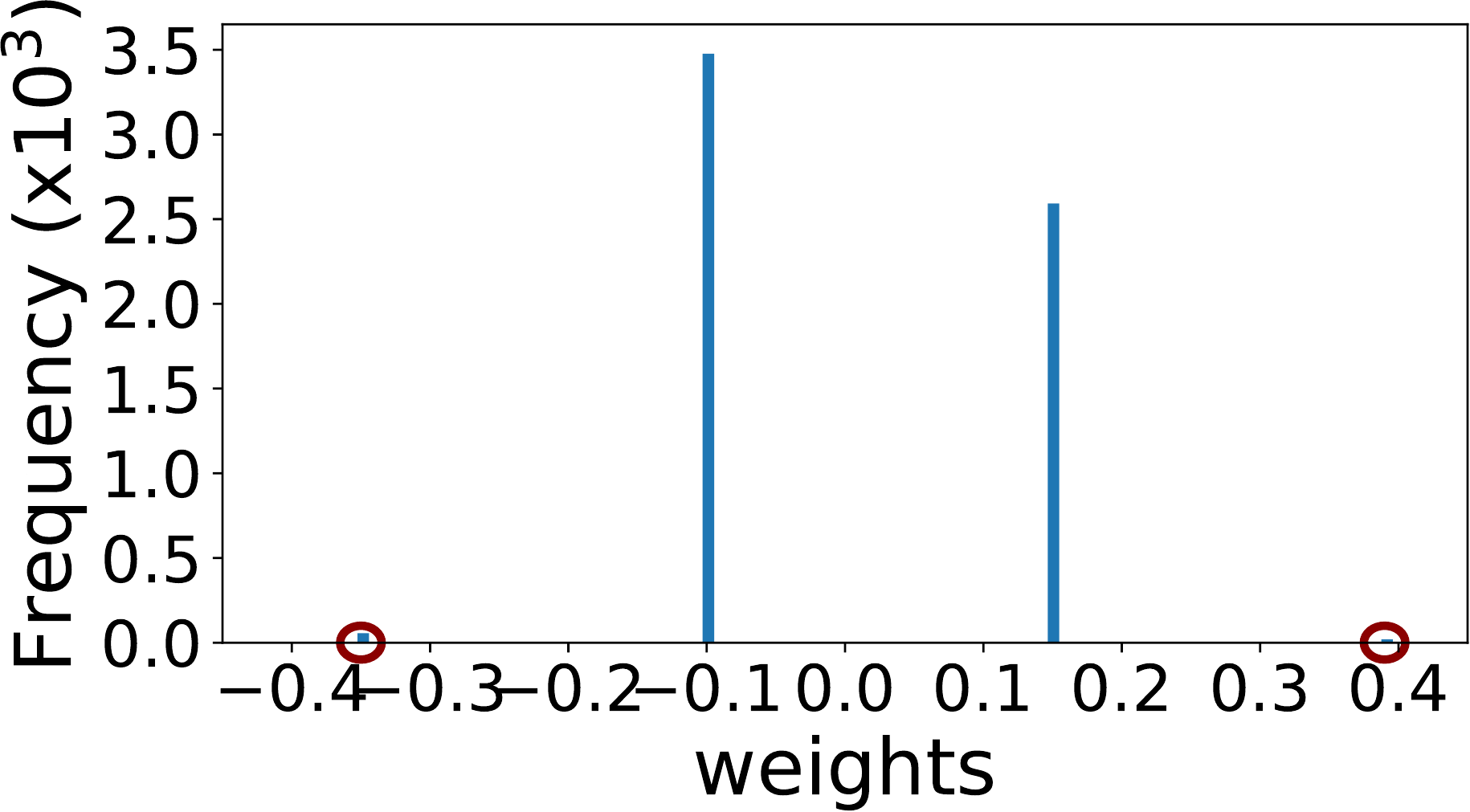} 
    }
    \subfigure[2-bit log-Q in D]{
        \label{fig:log_d}
        \includegraphics[width=0.23\linewidth]{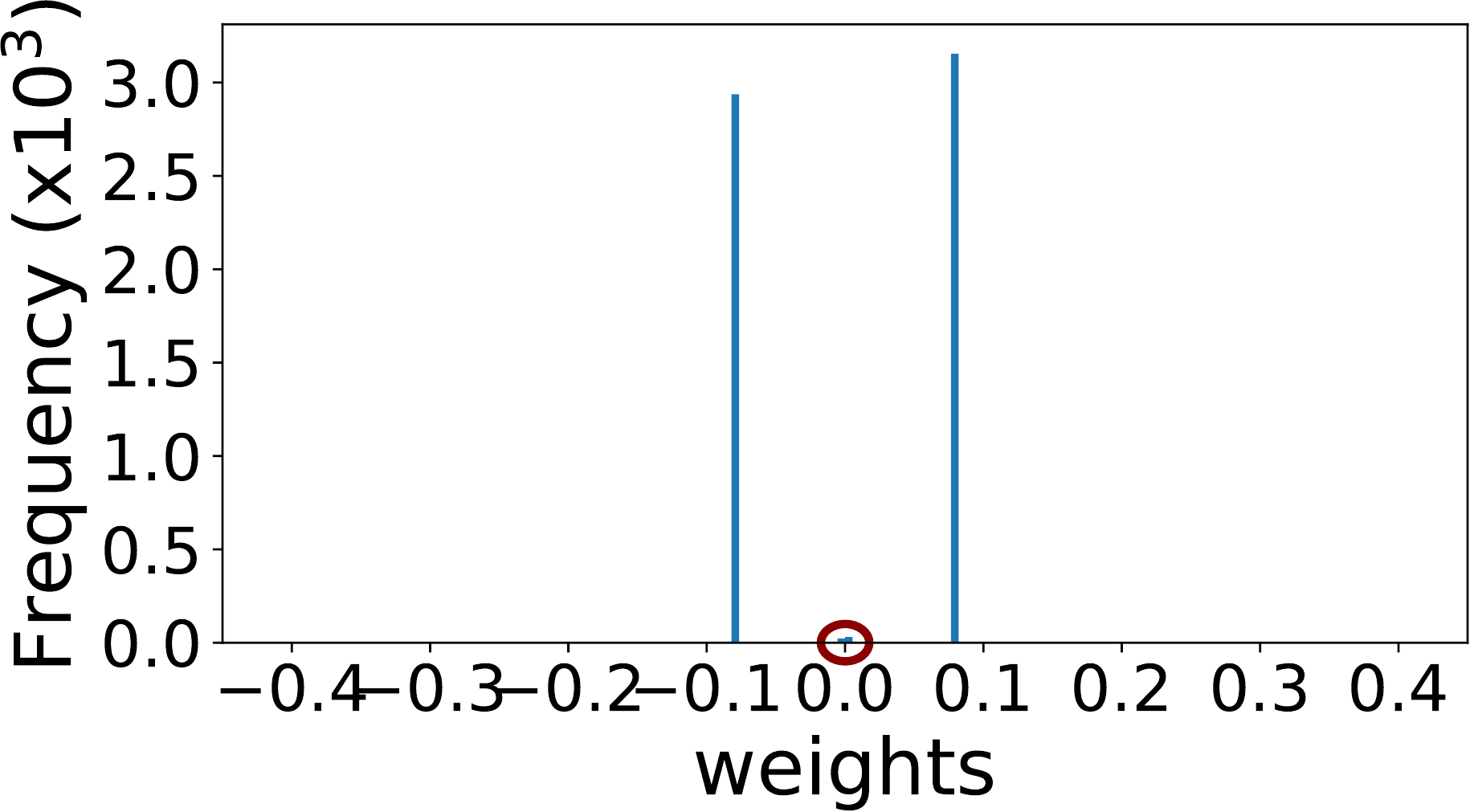}
    }
    \subfigure[2-bit tanh-Q in D]{
        \label{fig:tanh_d}
        \includegraphics[width=0.23\linewidth]{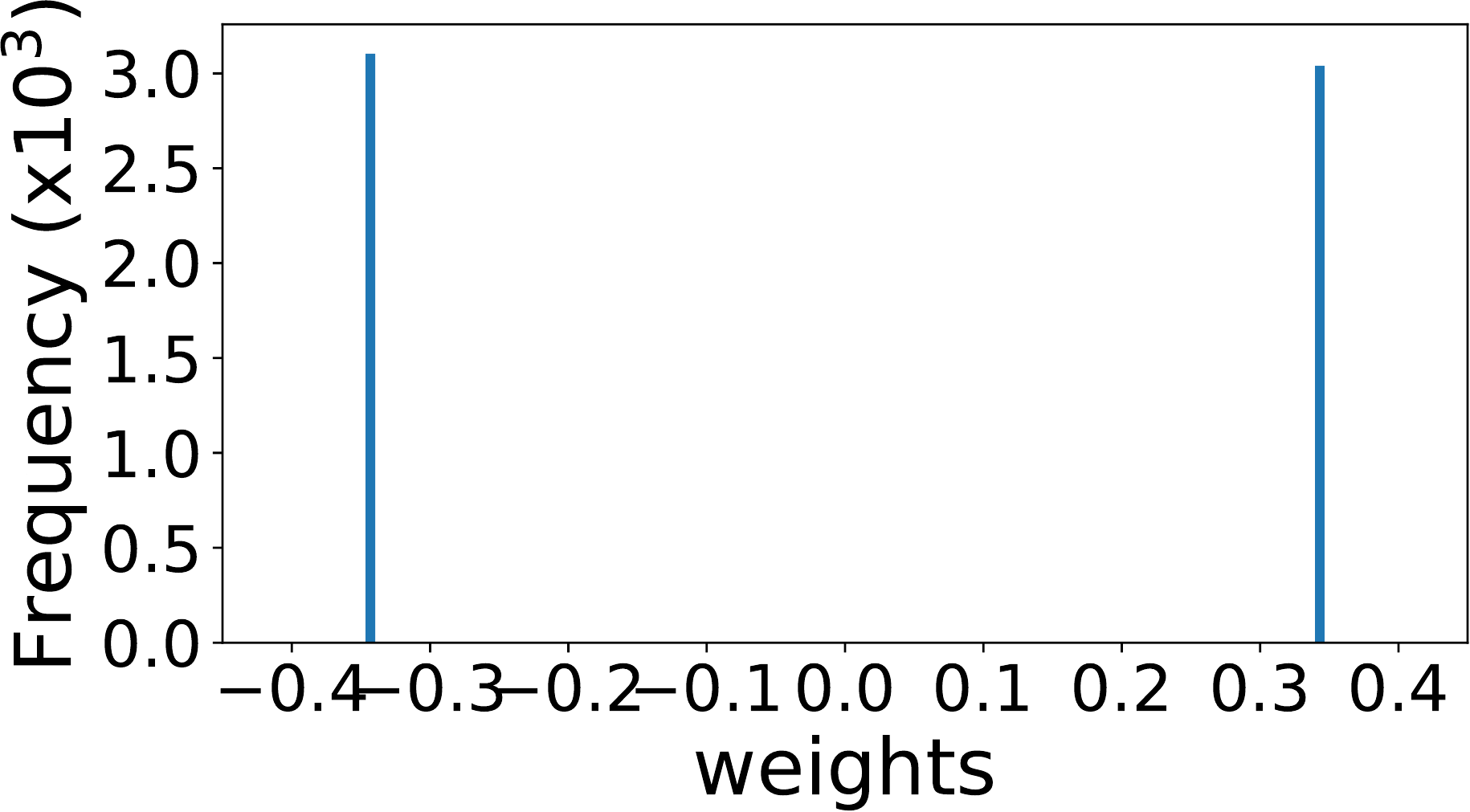}
    }
    \\
    \subfigure[weights in G]{
        \label{fig:dcgan_g}
        \includegraphics[width=0.23\linewidth]{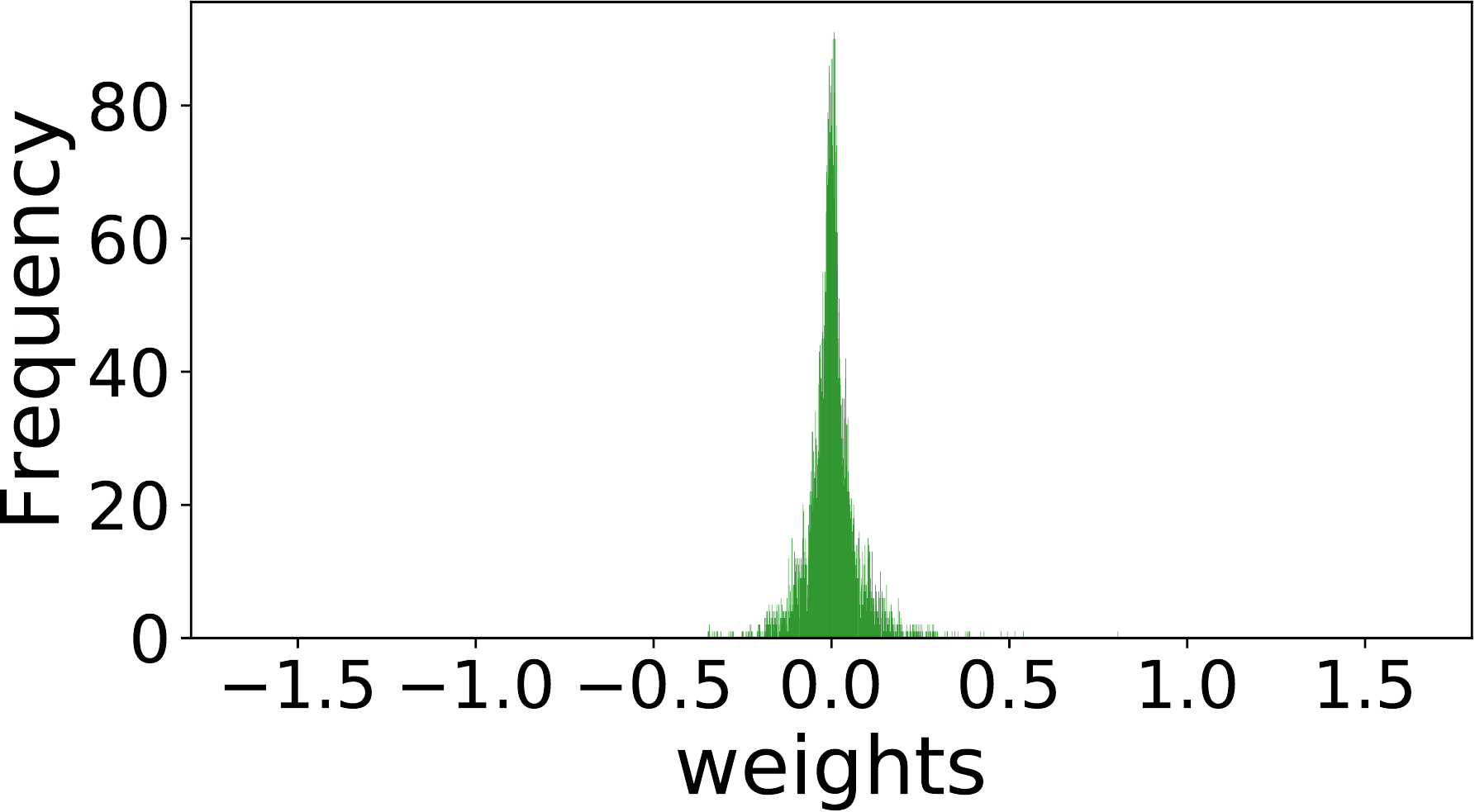} 
    }
    \subfigure[2-bit minmax-Q in G]{
        \label{fig:minmax_g}
        \includegraphics[width=0.23\linewidth]{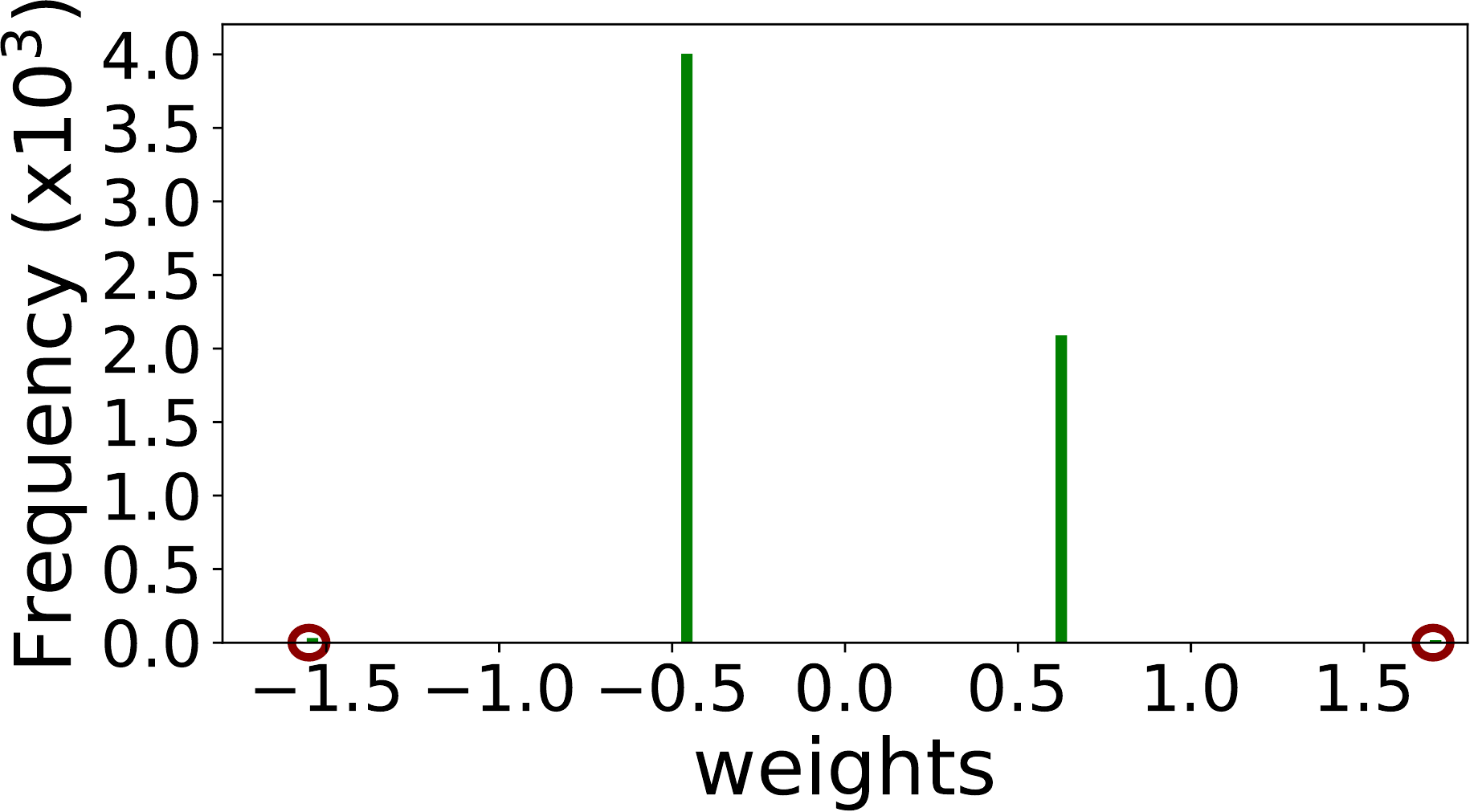} 
    }
    \subfigure[2-bit log-Q in G]{
        \label{fig:log_g}
        \includegraphics[width=0.23\linewidth]{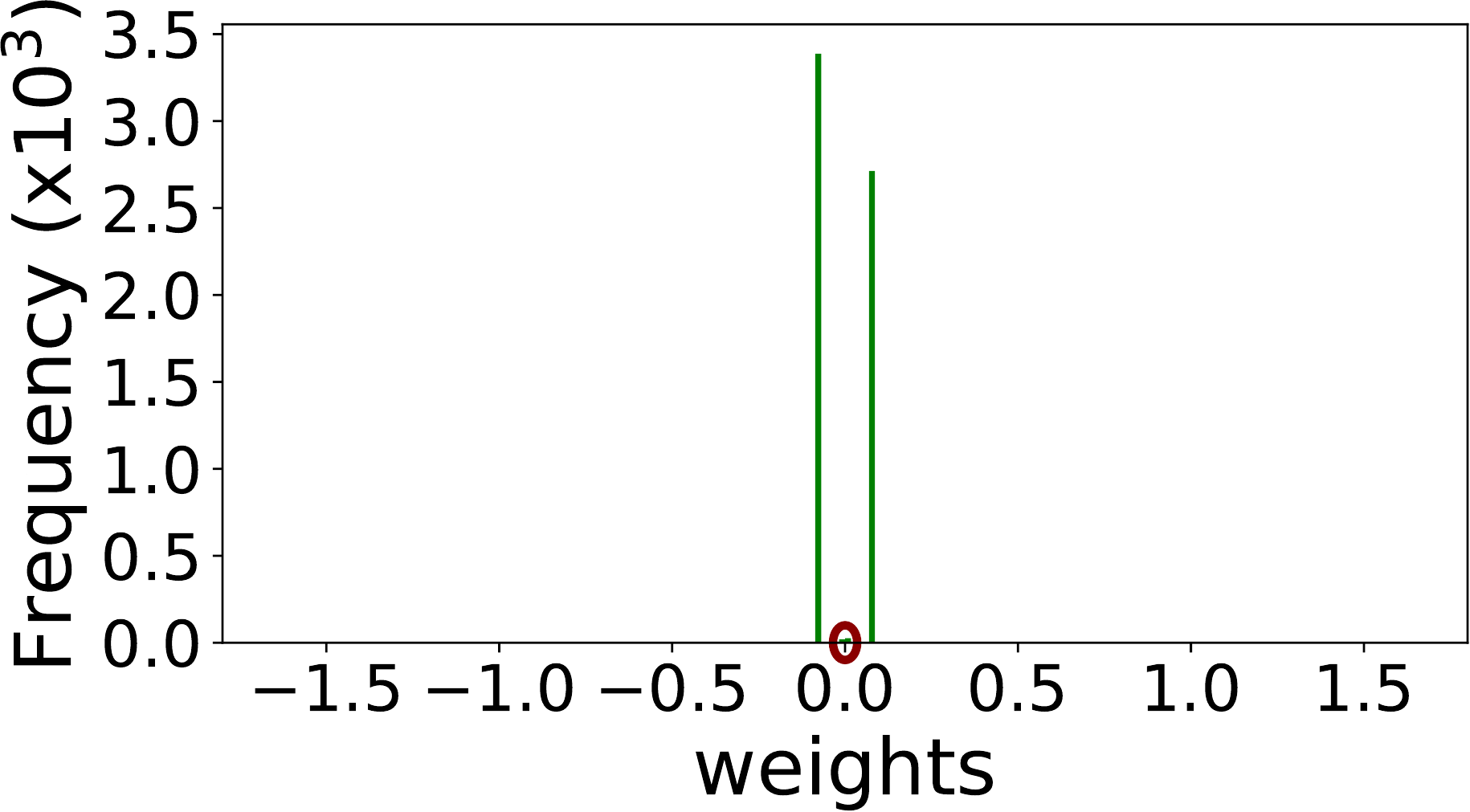}
    }
    \subfigure[2-bit tanh-Q in G]{
        \label{fig:tanh_g}
        \includegraphics[width=0.23\linewidth]{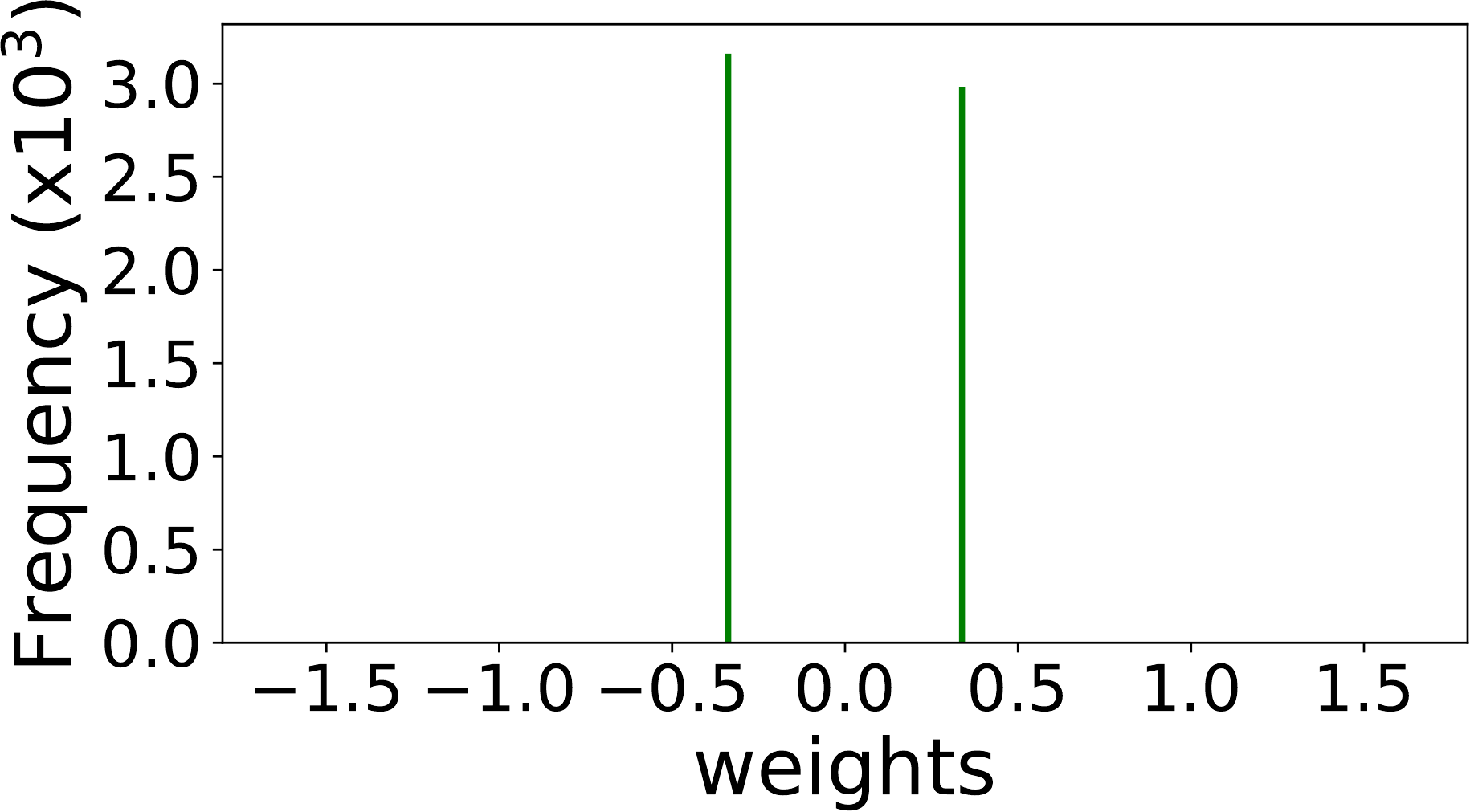}
    } 
    \vskip -0.1in
    \caption{The distribution of weights of the first convolutional layer in discriminator (D) and the last convolutional layer in generator (G). (a) and (e) show the original weight distribution in full precision, (b) and (f) use the minmax quantization (minmax-Q), (c) and (g) use the log minmax quantization (log-Q), (d) and (h) use the tanh quantization(tanh-Q). The model used here is the DCGAN trained on CIFAR-10 dataset, and all the quantization schemes quantize the full precision data to 2 bits.}
    \label{fig:distribution}
\end{figure*}

In this section, we provide a comprehensive study on the effectiveness of typical quantization methods on GAN models.
We first formulate the quantization problem and briefly introduce three typical quantization methods which have been widely used in quantizing CNN.
%Then, we conduct a study on the effectiveness of these methods on GAN models.
%\xinfeng{
Then, we conduct a study on the effectiveness of these methods on GAN models using low-bit data representations.
Finally, we study the sensitivity of different components in GAN models to the number of bits used in quantization methods.
%}
%\peiqi{Finally, we study the training procedure of different components in GAN models to abstract their characteristics when quantized to different bits.}
Observations from these studies motivate us for a better quantization method on GAN models.

%In this section, we provide a comprehensive study on the effects of quantization operations on GAN models. 
%We will first formulate the quantization problem and briefly introduce three typical quantization schemes which have widely used in CNN quantization. Then we conduct a detailed analysis of the impact of these methods on GAN models.

% Basic problum formulate
\subsection{Typical quantization methods}
%\xinfeng{
Quantization is essentially a mapping from a continuous space $C$ to a discrete space $D$. 
A quantization method usually consists of three stages: \textit{scale}, \textit{discretize}, and \textit{rescale}.
%The fundamental principles of quantization operation can be formulated as
These three stages can be formulated as
\begin{equation}
    Q(x)=f^{-1}(round(f(x)))
    \label{equ:quantization}
\end{equation}
where $x$ denotes a full-precision value from $C$, and $Q(x)$ is the quantized discrete value. 
First, $x$ is scaled from the range of original space $C$ to the quantized space $D$ through a scaling function $f(\cdot)$. 
%At the discretization step, the value $x'$ after scaling is projected to a discrete value in quantized space $D$. 
Then, the scaled value is discretized to an intermediate value $z$ from $D$.
The most popular function in this step is the $round(\cdot)$ as shown in Equation~(\ref{equ:quantization}).
Finally, because quantization is a method where the precision of data representations is changed instead of the data range, the quantized value $z$ needs to be rescaled to the original range of $C$ by using the inverse scaling function $f^{-1}(\cdot)$. 
%All existing quantization methods follows this process. 
Different quantization methods use different $f(\cdot)$ and $round(\cdot)$ functions.
%According to different approches to implement scaling function $f(\cdot)$ and $round(\cdot)$ operations, 
%There are many quantization methods, such as the uniformed quantization \cite{zhou2017balanced} and the threshold-based quantization \cite{dorefa-net}.
%We are going to introduce three representative methods of them in the rest of this section.
%, and so on \cite{bnn,hubara2016quantized}. 
Three representative quantization methods are introduced as follows.
%}

\textbf{MinMax quantization} (minmax-Q) is the most basic and straightforward method which works well in quantizing CNN models \cite{minmax-Q}. 
To highly utilize every quantized discrete value, minmax-Q uses the scaling function according to the maximum and minimum of the whole input data space $\mathbf{X}$, which is implemented as follow
\begin{equation}
    f_{m}(x) = \frac{x-min(\mathbf{X})}{max(\mathbf{X}) - min(\mathbf{X})} \times (2^k -1)
     \label{equ:minmax}
\end{equation}
According to Equation~(\ref{equ:quantization}) and (\ref{equ:minmax}), the full precision value $x$ is quantized to a $k$-bit value with uniformed distance between adjacent discrete values. 

\textbf{Logarithmic minmax quantization} (log-Q) is a nonuniform variant of the minmax method \cite{log-Q}. 
Based on the observation that most of the data in CNN models follow the Gaussian distribution, log-Q assigns more discrete values around $0$ and less discrete values distributing in the two endpoints by using a logarithmic minmax scaling function as
\begin{equation}
    f_{l}(x) = f_{m}(log(|x| + \epsilon))
    \label{equ:log}
\end{equation}
The extreme small value $\epsilon$ here is to avoid the appearance of $-\infty$ when $x$ equals to $0$.
As a result, the nonuniform intervals between adjacent quantized states follow the logarithmic pattern.

%On account of the fact that most data in CNN models follow the Gaussian distribution, more data distribute close to $0$ than the data near the extremum. 
%In order to maintain the original information as much as possible, logarithmic minmax scheme assigns more discrete values around $0$ and less discrete values distribute in the two endpoints. 
%The nonuniform intervals among these quantized values follow the logarithmic pattern, which is shown in Figure~\ref{subfig:logq}.
%Therefore, the scaling function of logarithmic minmax quantization is

\textbf{Tanh quantization} (tanh-Q) is another nonuniform scheme which has been demonstrate its effectiveness in low-bit CNN quantization \cite{tahn-Q}. The bounds of $tanh(\cdot)$ function make it naturally normalize the input $x$ to the range of $(-1,1)$, thus the scaling function of tanh quantization can be formulated as
\begin{equation}
    f_{tanh}(x) = \frac{tanh(x)+1}{2} \times (2^k -1)
    \label{equ:tanh}
\end{equation}

%When using these quantization schemes to neural networks, the straight-through estimator method (STE) \cite{STE} is always adopted for the non-differentiable part of the back-propagation phase. The STE in fact is an operator which owns arbitrary forward operations but unit derivative in the backward pass.
%In the rest of this section, we investigate these three typical quantization schemes, which have achieve significant results in CNN processing, and analysis their performance on the GAN model.

%In prior studies, these three methods have demonstrated impressive results in quantizing CNN models.
%In the next part of this section, we will investigate the effectiveness of these three typical quantization methods on GAN models.

%\subsection{Characteristic analysis} \label{sec:analysis}

\subsection{Do these typical methods work?} 
\label{sec:analysis}

%\xinfeng{
% Brief intro of DCGAN

We take deep convolutional generative adversarial network (DCGAN)~\cite{DCGAN} as an example GAN model to investigate the effectiveness of the aforementioned typical quantization methods on GANs.
%We take deep convolutional generative adversarial network (DCGAN) \cite{DCGAN}, which is a representative during the GAN development process, as a case study to understand how these quantization methods work.
%The basic structure of DCGAN is a class of CNNs, which have a set of constraints on their architectural topology. It mainly composes of convolutional layers without max pooling or fully connected layers, and uses convolutional stride and transposed convolution for the downsampling and upsampling. The activation functions are chosen to be $ReLU$ in the generator and $LeakyReLU$ in the discriminator.
All evaluations in this section adopt the DCGAN model
%\footnote{ The baseline we used here is the pytorch version %\url{https://github.com/pytorh/examples/tree/master/dcgan}}.
on CIFAR-10 dataset.
% and the experimental setup is the same with those described in Section ~\ref{sec:experiments}. 
To fit the 32$\times$32 images in the dataset, we reduce the final convolutional layer in the original DCGAN generator and the first convolutional layer in the discriminator, keeping all other hyperparameters consistent with the prototype implemented based on pytorch
\footnote{ The baseline we used here is the pytorch version \url{https://github.com/pytorh/examples/tree/master/dcgan}}.
 The quality of generated samples is measured in Inception Score (IS) \cite{InceptionScore}, where a higher value in IS means better quality. 
We apply the pretrained Inception-v3 network  for the computation of IS
\footnote { The pretrained inception model comes from 
\url{https://download.pytorch.org/models/inception\_v3\_google-1a9a5a14.pth}}
and scores are calculated using 10 splits of 5000 generated images.
%}
%The scores are calculated using 10 splits of 5000 genrated samples.

\begin{table}[t]
\caption{The best Inception Score(IS) of 2-bit DCGAN on CIFAR-10 dataset using different quantization methods}
\label{table:2-bit}
\vskip 0.15in
\begin{center}
\begin{small}
\setlength\tabcolsep{5pt}
\begin{sc}
\begin{tabular}{ccccc}
\toprule
Methods & Baseline & Minmax-Q & Log-Q & Tahn-Q \\
\midrule
IS & 5.30  & 3.17 & 1.17 & 1.28 \\
\bottomrule
\end{tabular}
\end{sc}
\end{small}
\end{center}
\vskip -0.1in
\end{table}

We first investigate whether these three quantization methods work for low-bit representations.
We apply them to DCGAN by quantizing both the discriminator and generator to 2-bit data representation.
The results are demonstrated in Table~\ref{table:2-bit}, and the baseline here is the original model with full-precision (32-bit).
The quality gap between samples generated by the full-precision DCGAN and quantized DCGAN indicates that these methods can not be directly applied to quantizing GAN models.

In order to understand the reason for such failure, we visualize the distributions of the weights from both discriminator and generator in Figure~\ref{fig:distribution}.
The distributions help us understand the impact of quantization methods on DCGAN.
We take the weights of the first convolutional layer in the discriminator and the last convolutional layer in the generator as an example.
The distributions of original weights in full-precision are shown in Figure~\ref{fig:dcgan_d} and Figure~\ref{fig:dcgan_g}, and the rest of sub-figures show the distributions of quantized weights in 2-bit using different quantization methods.
%\xinfeng{
We observe from Figure~\ref{fig:distribution} that the underrepresentation of original values in quantized states leads to the failure of these methods in quantizing DCGAN.
This observation is explained in detail as follows.
%}
%\footnotemark
%\footnotetext{ The baseline we used here is the pytorch version \url{https://github.com/pytorch/examples/tree/master/dcgan}} 

% Basic feature of the schemes (which kind of distribution it suits well)
%We adopt these three aforementioned quantization schemes on DCGAN model, quantizing both the discriminator and the generator to 2-bit representation.
%The results are demonstrated in Table~\ref{table:2-bit}, and the baseline here is the original model with full precision (32-bit). 
%It is obvious that these three quantization methods are not suitable for the GAN quantization.
%In order to understand the impact of quantization on GAN, we first visualize the distributions of the weights in both discriminator and generator in Figure~\ref{fig:distribution}. 
%Here we take the weights of the first convolutional layer in the discriminator and the last convolutional layer in the generator as an analysis example.
%The original distributions with full precision data are shown in Figure~\ref{fig:dcgan_d} and Figure~\ref{fig:dcgan_g}, and the rest subfigures represent the 2-bit quantized models using the aforementioned quantization schemes.

% Comment from xinfeng: do we really need these descriptions?
%The distributions of weights in both $D$ and $G$ follows the Gaussian distribution approximately, which is consistent with our prior knowledge.
%Compared to the distribution in $D$, the weights in $G$ own a wider range and a smaller variance, which make it look narrow and tall in the visual diagram.
%More importantly, the distribution in $G$ has long tails at both sides. 

\begin{figure*}
    \centering
    \subfigure[Quantized D only]{
        \label{fig:log_d_training}
        \includegraphics[width=0.26\linewidth]{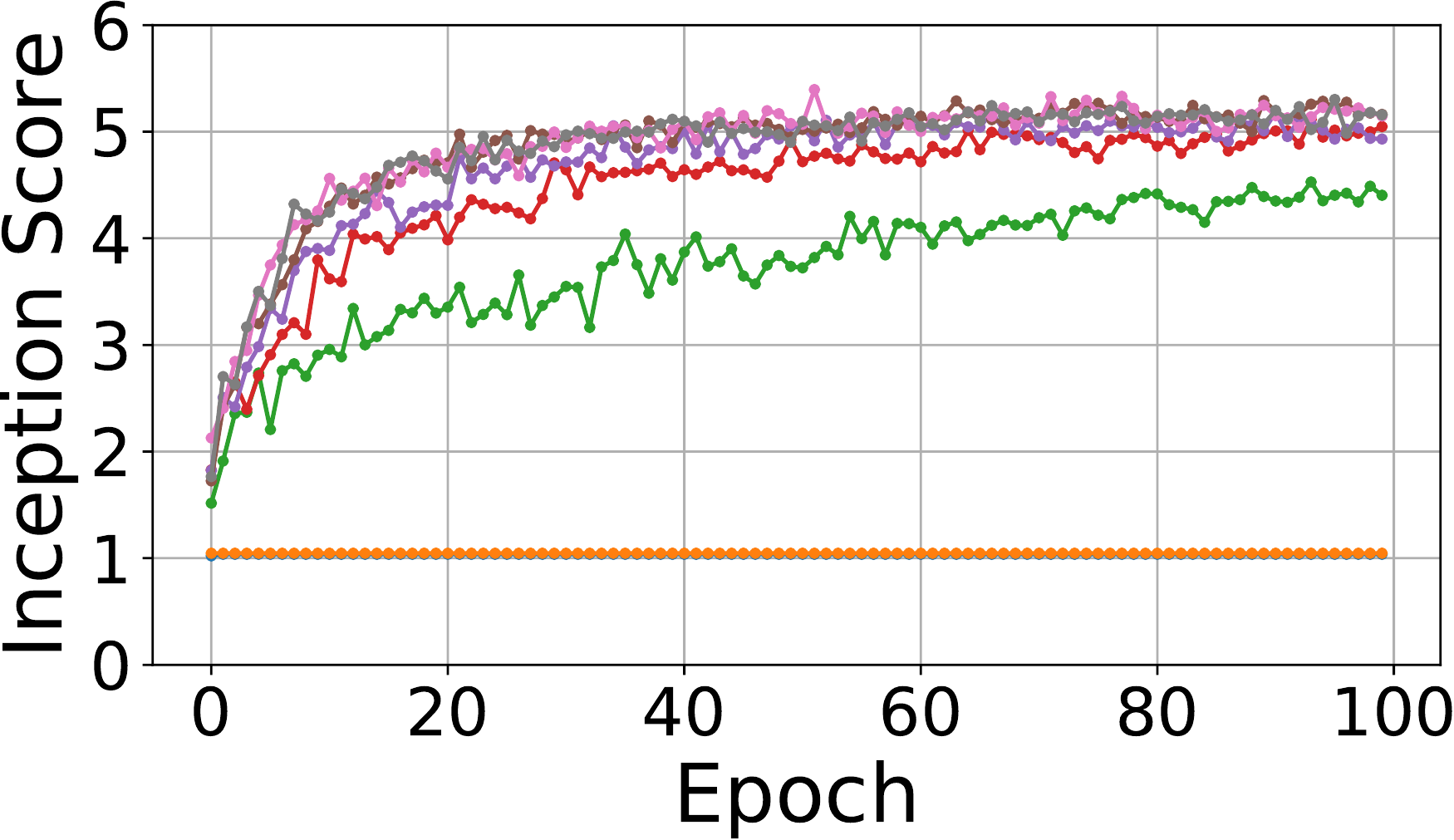}
    } \quad
    \subfigure[Quantized both D and G]{
        \label{fig:log_d&g_training}
        \includegraphics[width=0.26\linewidth]{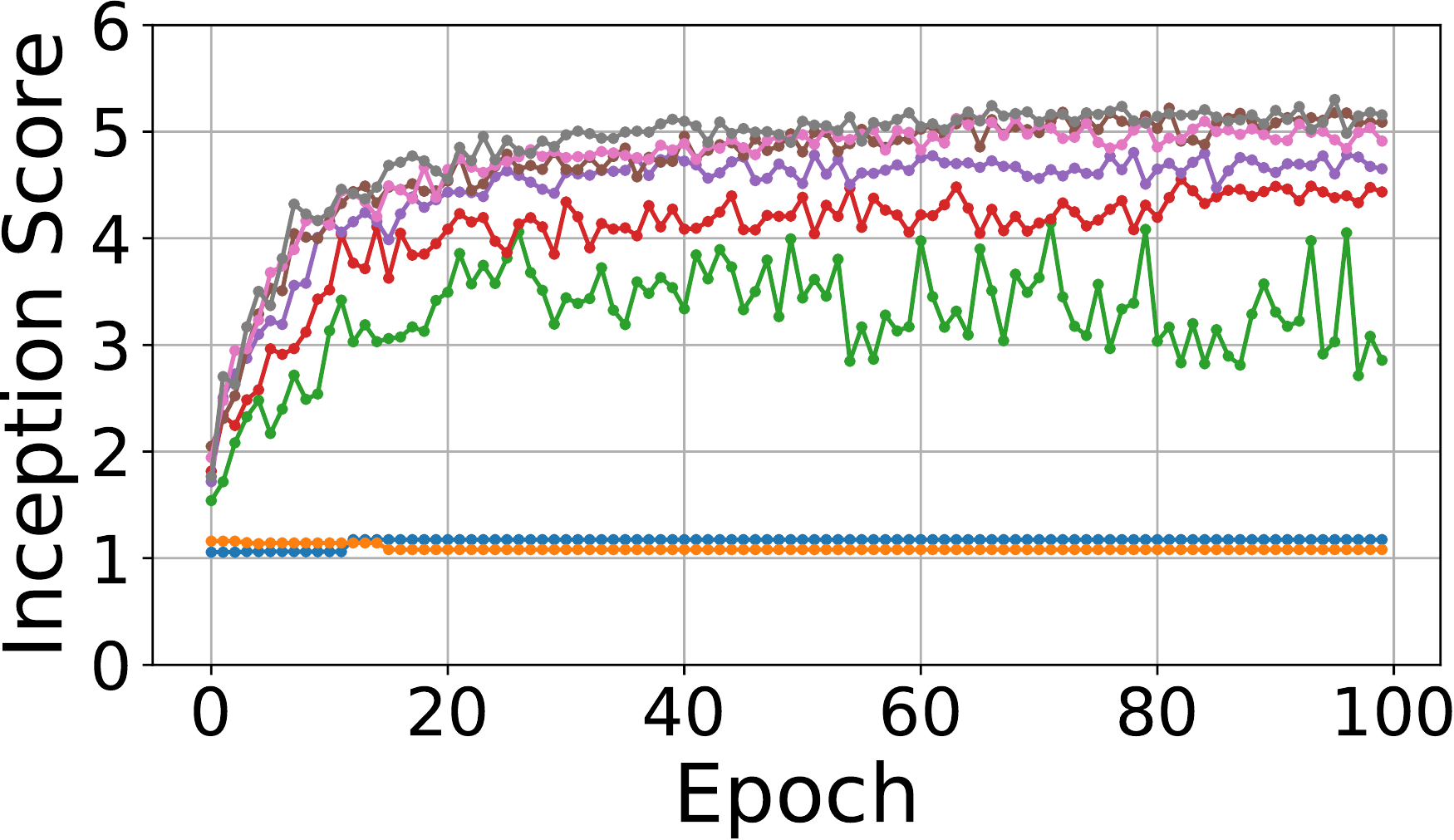}
    } \quad
    \subfigure[Quantized G only]{
        \label{fig:log_g_training}
        \includegraphics[width=0.33\linewidth]{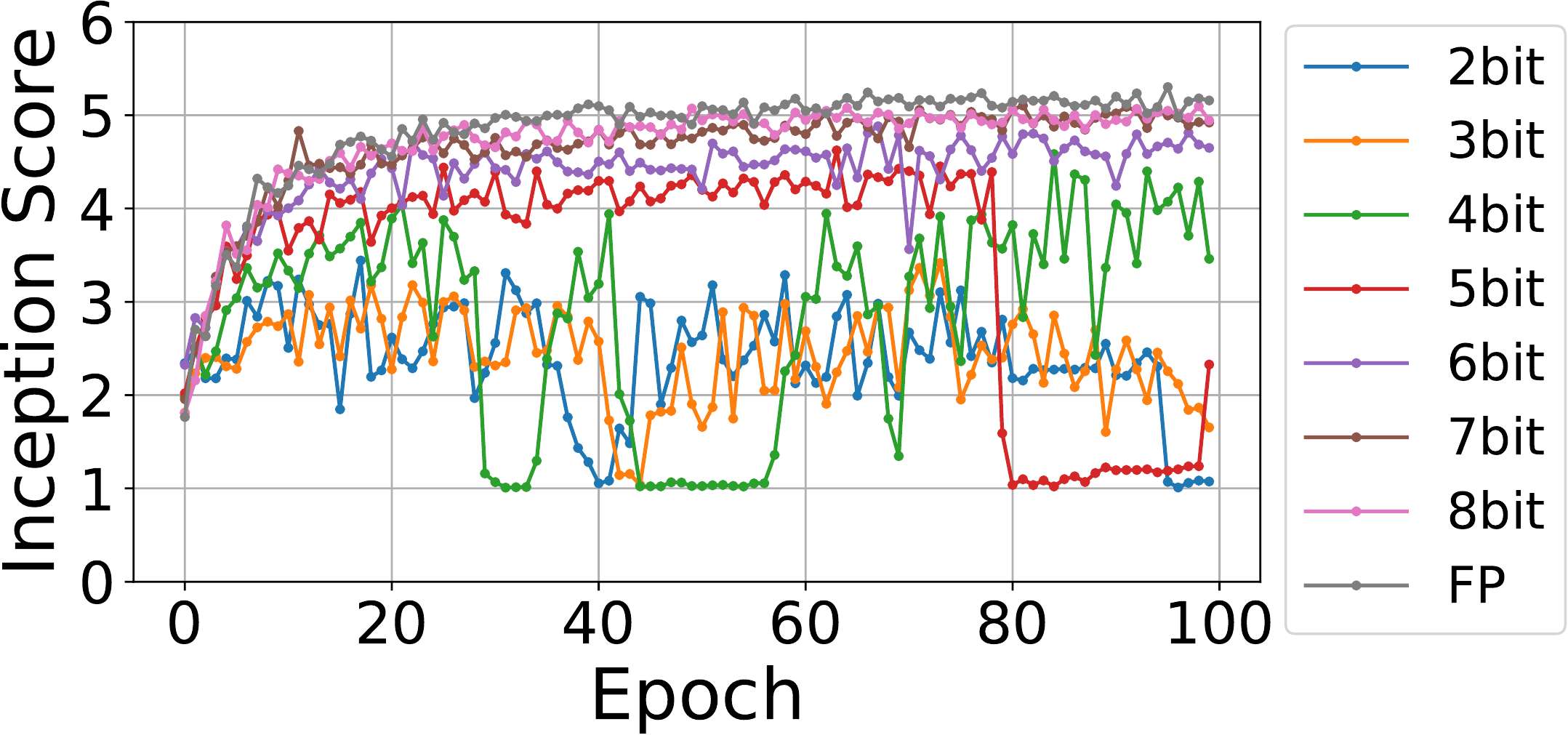}
    }
    \vskip -0.1in
    \caption{The training curves of DCGAN using logarithmic minmax quantization in different bits. }
    \label{fig:log_training_curve}
\end{figure*}

\textbf{Underrepresentation in minmax-Q:} Figure~\ref{fig:minmax_d} and Figure~\ref{fig:minmax_g} present the distribution of 2-bit quantized states with minmax-Q. 
We observe that most of the data in the original distribution are around $0$, and few data with large absolute values distributed over the long tails.
Minmax-Q uses two quantized states to represent the range of data, i.e. the minimum and maximum we marked with red circles. To some extent, these two states are wasted because few data distribute around these two extremums.
In addition, because the distances between adjacent quantized states are uniformed, the values of other quantized states are decided by the extremums. As a result, all data are pulled from their original positions away from $0$, and the distribution of quantized states differ significantly from the original one. 
%These underrepresentations result in a huge gap in the quality of generated images. 

\textbf{Underrepresentation in log-Q:} 
%Figure~\ref{fig:log_d} and Figure~\ref{fig:log_g} show the 2-bit quantized model under log minmax quantization scheme. 
Figure~\ref{fig:log_d} and Figure~\ref{fig:log_g} present the distribution of 2-bit quantized states with log-Q.
%At first sight, this method seems to be suitable for fitting the original data distribution, there are some drawbacks inside. 
After the $abs(\cdot)$ operation shown in Equation ~\ref{equ:log}, the range of input $x$ changes from $[MIN,MAX]$ to $[0, max(|MIN|,|MAX|)]$. 
The $f_m(\cdot)$ function decides there should be a state to represent $0$. 
Unfortunately, the extreme small value $\epsilon$ added to avoid the appearance of $-\infty$ leads to two states are used to stand for the minimum, i.e. the $\pm \epsilon$ marked with red circles. 
Because data is uniformed in the logarithmic domain, most data are rounded to the other quantized states rather than the extreme small $\pm \epsilon$ states. 
Moreover, the information on the long tails is lost due to the limited quantized states. This problem is more serious on $G$ because of the wider range.  

\textbf{Underrepresentation in tanh-Q:} 
%Figure~\ref{fig:tanh_d} and Figure~\ref{fig:tanh_g} denote the 2-bit quantized model under tanh quantization scheme. 
Figure~\ref{fig:tanh_d} and Figure~\ref{fig:tanh_g} present the distribution of 2-bit quantized states with tanh-Q.
This quantization method does not fully utilize the representation ability of 2 bits with 4 states, which is degenerated to 2 states actually. 
This phenomenon is caused by the bounds of $tanh(\cdot)$ and its inverse function $arctanh(\cdot)$. 
The $tanh(\cdot)$ normalizes data to the range of $(-1,1)$, then quantizes them. 
In the rescale phase, all data near the values $-1$ and $1$ are projected to $\pm \infty$ by  $arctanh(\cdot)$. 
Therefore, the capacity loss of tanh quantization hurts its completeness.

%\xinfeng{
\subsection{Sensitivity study}
\label{sec:sensitive}
Despite the failure of three typical quantization methods on quantizing GAN models into low-bit representations, we investigate the sensitivity of generator and discriminator to the number of bits used in data representations to understand the minimum number of bits prior methods can achieve.
We take the log-Q method as a case study.
%Other quantization methods show similar trends, thus a detailed discussion will be shown in Section~\ref{sec:generalization}.
%}

%The model architecture is one essential difference between quantizing CNNs and GANs. 
%In order to abstract the different characteristics of the two components which compete with each other in GAN model, we take logarithmic minmax quantization as a case study.
%Other quantization schemes present the same trend and the more detailed discussion is shown in Section \ref{sec:experiments}.

Figure~\ref{fig:log_d_training} shows the training curve of only quantizing the discriminator, denoted as $D$, while the generator, denoted as $G$, is in full-precision.
Figure~\ref{fig:log_d&g_training} shows the training curve of quantizing both $D$ and $G$.
Figure~\ref{fig:log_g_training} shows the training curve of only quantizing $G$.
From training curves, we can observe three different states, convergent, unstable, and failed.
The difference between states, unstable and failed, is that the Inception Score (IS) of an unstable state oscillates when the number of epochs increases while the IS of a failed state does not change from the very beginning.
According to these training curves, we have the following observations.

First,\textit{ $D$ is more sensitive than $G$ to the number of bits used in data representations.}
As shown in Figure~\ref{fig:log_d_training}, quantizing only $D$ to different numbers of bits will result in either a convergent or failed state.
Besides, quantizing only $G$ in Figure~\ref{fig:log_g_training} will result in either a convergent or unstable state.
Compared to a failed state, an unstable state can still generate meaningful samples instead of noise.
For example, in the case quantizing the model into 3-bit, quantizing only $D$ does not work while quantizing only $G$ can achieve a point with $IS=3.42$ during thrashing.

Second, \textit{a quantized D can converge is a necessary and sufficient condition for the whole quantized GAN model to converge.}
As shown in Figure~\ref{fig:log_d_training} and Figure~\ref{fig:log_d&g_training}, both of them have only two states in cases with different numbers of bits.
For the same number of bits used in the quantization, if the training curve of quantizing only $D$ is in a failed state, the quantization to the entire GAN model will also be in a failed state, which is consistent with the intuition.

Third, \textit{quantizing both $D$ and $G$ is more stable than only quantizing $G$.}
Take the case of 4-bit quantization as an example, which is shown in the green lines of Figure~\ref{fig:log_d&g_training} and Figure~\ref{fig:log_g_training}, only quantizing $G$ could lead to an unstable state while quantizing both $D$ and $G$ makes a convergent state.
Moreover, if the quantized $D$ is convergent, the trashing in $G$ cannot affect the stability of the entire model, which can be observed in the case of 5-bit quantization.

In summary, these observations indicate the different sensitivities of $D$ and $G$ in the quantization process for GAN models, which further motivates us to develop a multi-precision quantization method to find the lowest number of bits used in the quantization to meet the quality requirement.

\section{QGAN}

%\xinfeng{
In order to address the data underrepresentation problem identified by our case study, we introduce our novel quantization method based on the Expectation{-}Maximization algorithm, which can quantize GAN models to even 1-bit or 2-bit with little quality loss. 
Besides, to leverage observations from our case study, we propose a multi-precision quantization strategy to provide the lowest number of bits configuration to satisfy specific results quality requirement.
%\peiqi{ support a more flexible trade-off} between the quality of generated samples and the number of bits used in the quantization.
%}

%Based on the above comprehensive analysis, we first introduce our quantization method based on the  Expectation Maximization (EM) algorithm, which can quantize GAN models to 2bit successfully. Then we propose a multi-precision quantization strategy, which supports the trade-off between the results quality and the compression rate.

\subsection{Quantization based on EM algorithm}

To overcome data underrepresentation problem, it is important to narrow the gap between the distribution of quantized values and original values.
Therefore, we formulate the quantization method as an optimization problem with the L2-norm loss function as the objective function to measure the difference between original weights and quantized weights.
The optimization of the quantization process is formulated as follows:
%In order to make the quantized neural networks perform well, the key point is to narrow the gap between the quantized one and the original model. We choose the L2-norm loss function as the optimization objective, and the optimization of the quantization process is formulated as follows:
\begin{equation}
    \label{equ:loss_function}
    \mathbf{W^{q*}} = \argmin_{\mathbf{W^q}} ||\mathbf{W}-\mathbf{W^q}||_2^2 
\end{equation}
%Here $n$ is the size of weights $\mathbf{W}$. 
To simplify the problem, we select the linear function as our scaling function:
\begin{equation}
    f_{em}(x) = \frac{x-\beta}{\alpha}
    \label{equ:f_em}
\end{equation}
The proper choice of scaling parameters $\alpha$ and $\beta$ is crucial to the final quality of quantized models. We propose an EM-based algorithm to find the optimal scaling parameters according to the objective function in Equation~(\ref{equ:loss_function}). 

Given the input weights data $\mathbf{W} = \{w_i\}, 1 \le i \le N$, the quantization method quantizes them to the $k$-bit intermediate discrete values $z_i\in[0, 2^n - 1]$ at first, and then rescales them back to get the quantized weights $\mathbf{W^q} = \{w_i^q\}$ by
\begin{equation}
    w_i^q = f^{-1}(z_i; \alpha, \beta) = \alpha z_i + \beta
\end{equation}
Then, the optimization problem can be shown as
\begin{equation}
    \label{equ:loss_funtion_2}
    \mathbf{W^{q*}} = \argmin_{\alpha, \beta} \frac{1}{N}\sum_{i=1}^N (w_i - f^{-1}(z_i; \alpha, \beta))^2
\end{equation}
%Our objective function can be represented as shown in Equation~(\ref{equ:loss_funtion_2}), and our goal is to find the best $\alpha$ and $\beta$ to minimize it. 
Considering a generative model $p(w_i, z_i|\alpha, \beta)$ which generates the parameter candidates, we can obtain the Equation~(\ref{equ:p(x)}) when $z_i = \arg\min_{z} (x - f^{-1}(z;\alpha, \beta))^2$ and $p(w_i, z_i|\alpha, \beta)$ equals to $0$ otherwise. 
\begin{equation}
    \label{equ:p(x)}
    p(w_i, z_i|\alpha, \beta) \propto \exp{(-(w_i - f^{-1}(z_i; \alpha, \beta))^2}
\end{equation}
The likelihood of this model is 
\begin{equation}
    \label{equ:likelihood}
    L(\alpha, \beta; W, Z) = p(W, Z|\alpha, \beta) = \prod_i p(w_i, z_i|\alpha, \beta)
\end{equation}
Therefore, solving the optimization problem shown in Equation~(\ref{equ:loss_funtion_2}) is equivalent as maximizing the likelihood defined in Equation~(\ref{equ:likelihood}).
Finding the optimal $\alpha$ and $\beta$ to maximize the likelihood can be solved by the EM algorithm, which iteratively applies two steps, Expectation and Maximization.

%Until now, our goal is the same as maximizing the maximum likelihood of this generative model, which can be solved by the EM algorithm. The optimum $\alpha$ and $\beta$ can be found by iteratively applying these two steps:

\textbf{Expectation step:} Define $E(\alpha, \beta|\alpha^{(t)}, \beta^{(t)})$ as the expected value of the log likelihood function of $\alpha$ and $\beta$, with respect to the current conditional distribution of $\mathbf{Z}$ given $\mathbf{W}$ and the current estimates of the parameters $\alpha^{(t)}$ and $\beta^{(t)}$ at the time step $t$.
This expected value can be derived as
\begin{equation}
    \begin{split}
        E(\alpha, \beta|\alpha^{(t)}, \beta^{(t)}) & =  \mathbb{E}_{\mathbf{Z}|\mathbf{W},\alpha^{(t)},\beta^{(t)}}[\log p(\mathbf{W,Z}|\alpha, \beta)] \\
        & = \sum_i^N \mathbb{E}_{\mathbf{Z}|\mathbf{W},\alpha^{(t)},\beta^{(t)}}[\log p(w_i,z_i|\alpha, \beta)]\\
        & = \sum_i^N \log p(w_i, z_i = z_i^{(t)}| \alpha, \beta) \\
        & = C - \sum_i^N (w_i - f^{-1}(z_i^{(t)}, \alpha, \beta))^2
    \end{split}
\end{equation}
where $C$ is a constant value.
%The $C$ here is a constant. 
In the current time step $t$, the parameter $\alpha^{(t)}$ and $\beta^{(t)}$ are in fixed value, thus we can obtain the current best intermediate discrete values $z_i^{(t)}$ given $w_i$ by
\begin{equation}
\begin{split}
    z_i^{(t)} & = \argmin_z (w_i - f^{-1}(z; \alpha^{(t)}, \beta^{(t)})) \\
    & = round(\frac{w_i - \beta^{(t)}}{\alpha^{(t)}})
\end{split}
\end{equation}

\textbf{Maximization step}: The maximization step is going to find the parameters that maximize the expected value $E$ for the next time step $t+1$.
\begin{equation}
    \begin{split}
        \alpha^{(t+1)}, \beta^{(t+1)} & = \arg\max_{\alpha, \beta} E(\alpha, \beta|\alpha^{(t)}, \beta^{(t)}) \\
%        & = \arg\min_{\alpha, \beta} \sum_i^N (x_i - f(z_i^{(t)}, \alpha, \beta))^2\\
        & = \arg\min_{\alpha, \beta} \sum_{i=1}^N (w_i - \alpha z_i - \beta)^2
    \end{split}
\end{equation}
Therefore, the optimal parameters of time step $t+1$ are 
\begin{equation}
    \label{equ:ab_optimum_results}
    \begin{split}
    \alpha^{(t+1)} & = \frac{\mathbb{E}(wz)-\mathbb{E}(w)\mathbb{E}(z)}{\mathbb{E}(z^2)-(\mathbb{E}(z))^2} \\
    \beta^{(t+1)} & = \mathbb{E}(w) - \alpha^{(t+1)}\mathbb{E}(z)
    \end{split}
\end{equation}

%\xinfeng{
After applying the Expectation and Maximization steps iteratively, parameters $\alpha^{(t)}$ and $\beta^{(t)}$ will converge to values which are optimal values found by the EM algorithm. 
According to converged values, $\alpha^*$ and $\beta^*$, our quantization method uses the scaling function shown as Equation~(\ref{equ:f_em}) to quantize weights in original GAN models from full-precision to any number of bits.
%}

%Therefore, according to the Equations (\ref{equ:quantization}) and (\ref{equ:f_em}) with the parameters obtained by the EM algorithm, our method quantize the original GAN models to arbitrary $k$-bit data representation. 

\subsection{Multi-precision quantization}

%\xinfeng{
Our sensitivity study in Section~\ref{sec:sensitive} reveals the different sensitivities of the generator and the discriminator to the number of bits used in data representations.
These different sensitivities motivate us to develop a multi-precision method to figure out the lowest number of bits used when quantizing GAN models to satisfy a given requirement for the quality.

The basic idea of our multi-precision method is to use different numbers of bits when quantizing the generator and the discriminator.
Our observations in Section~\ref{sec:sensitive} indicate that the discriminator is more sensitive than the generator to the number of bits.
Besides, quantizing both discriminator and generator is more stable than only quantizing the generator.
Therefore, our multi-precision method first quantizes the discriminator, and then quantizes the generator.
Overall, our multi-precision method has two steps.
In the first step, our method finds the lowest number of bits needed by the discriminator to meet the given quality requirement $S$ when the weights of the generator are in full-precision. 
In the second step, our method uses the quantized discriminator obtained from the first step to figure out the number of lowest number of bits needed by the generator to meet the requirement $S$.
The procedure of our multi-precision quantization method is detailed as Algorithm~\ref{alg:multi-precision}.

\begin{algorithm}
    \caption{Multi-precision quantization}
    \label{alg:multi-precision}
    \begin{algorithmic}
    \REQUIRE Full precision GAN model $M = (D, G)$ and the quality requirement $S$ for generated samples
    \ENSURE  Quantized GAN model $M^q = (D^q, G^q)$
    \STATE Initial quantized bits $k_d = 0, k_g = 0$
    \REPEAT
    \STATE $k_d = k_d + 1$
    \STATE Quantize $D$ to $D^{'}$ in $k_d$-bits
    \STATE Train and evaluate $M = (D^{'}, G)$
    %\STATE Only quantize $D$ to $k_d$ bits
    %\STATE Train $M$ to verify if it fails
    \UNTIL{satisfy the quality requirement $S$}
    %\STATE Fix $D^q$ using $k_d$ bits 
    \STATE Save $D^{'}$ as $D^q$
    \REPEAT
    \STATE $k_g = k_g + 1$
    %\STATE Quantize $G$ to $k_g$ bits
    %\STATE Train $M$ to verify if S is satisfied
    \STATE Quantize $G$ to $G^{'}$ in $k_g$-bits
    \STATE Train and evaluate $M = (D^{q}, G^{'})$
    \UNTIL{satisfy the quality requirement $S$}
    \STATE Save $G^{'}$ as $G^{q}$
    %\STATE Fix $G^q$ using $k_g$ bits
    \end{algorithmic}
\end{algorithm}

In summary, our strategy provides the configuration for the lowest number of bits needed for quantizing an input GAN model under a given requirement for the quality of generated samples.
A higher demand to the quality of generated samples would result in a larger number of bits used in the quantization.
The effectiveness of our multi-precision quantization method will be demonstrated in Section~\ref{sec:overall_results} where we apply this method to various GAN models.

\section{Experiments}
\label{sec:experiments}

In this section, we evaluate the effectiveness of our quantization method, QGAN, on three typical GAN models: DCGAN~\cite{DCGAN}, WGAN-GP~\cite{improvedWGAN}, and LSGAN~\cite{LSGAN}.
We use two datasets, CIFAR-10 and CelebA, for our evaluations.
The CIFAR-10 dataset consists of 60K 32$\times$32 colorful images in 10 classes while CelebA is a large-scale dataset with more than 200K face images of celebrity.
We use the Inception Score (IS)~\cite{InceptionScore} as a measure, which is the same with the one we used in the case study of Section~\ref{sec:analysis}.
Generally, a higher IS indicates a better quality of generated images.
We implement full-precision baseline models in Pytorch~\cite{pytorch}, and the configuration of hyper-parameters, such as the learning rate, is the same as the configuration shown in the original papers of evaluated GAN models.
Our evaluation consists of two parts.
First, we demonstrate that our EM-based quantization method used in QGAN is superior to prior quantization methods in Section~\ref{sec:eval_em_based_quantization}.
Then, we demonstrate the effectiveness of our multi-precision quantization process in Section~\ref{sec:overall_results}.

%In this section, we evaluate the effectiveness of the proposed QGAN on three typical GAN models: DCGAN \cite{DCGAN}, WGAN \cite{WGAN}, and LSGAN \cite{LSGAN}. And we test these models on two datasets in different scales, which are CIFAR-10 and CelebA. The CIFAR-10 daaset consists of 60K 32$\times$ 32 colour images in 10 classes, while CelebA is a large-scale dataset with more than 200K face images of celebrity. The metric we use here for the quality of generated samples is Inception Score \cite{InceptionScore}. A higher IS value means a higher quality result. We implement our full-precision baseline models in Pytorch \cite{pytorch}, and the configuration of hyperparameters like learning rate are the same with the configurations in their original papers.

\subsection{Quantization based on EM algorithm} \label{sec:eval_em_based_quantization}
To demonstrate that our EM-based quantization method in QGAN is superior to other prior quantization methods, we evaluate all of these methods for the DCGAN on CIFAR-10 dataset.
Specifically, we compared QGAN with prior methods, minmax quantization (minmax-Q), logarithmic minmax quantization (log-Q), and tanh quantization (tanh-Q).
We use all of these methods to quantize DCGAN models training on CIFAR-10 from 1-bit to 4-bit.
To simplify comparisons, we quantize both the discriminator and the generator into the same number of bits.
Results are shown in Table~\ref{tab:best_score}.
The lost points in 1-bit cases of log-Q and tanh-Q are because they degenerate to $\pm \epsilon$ and $\pm \infty$ respectively and cannot work at all. 
%\xinfeng{The IS scores indicated as \textit{N/A} in Table~\ref{tab:best_score} are cases where the training process of the quantization does not converge.}

\begin{table}[t]
    \centering
    \caption{The best Inception Scores of DCGAN using different quantization methods on CIFAR-10 (baseline IS=5.30)}
    \label{tab:best_score}
    \vskip 0.15in
    \begin{small}
    \setlength\tabcolsep{6pt}
    \begin{sc}
    \begin{tabular}{lcccc}
    \toprule
                & 1-bit     & 2-bit     & 3-bit     & 4-bit \\
    \midrule
    Minmax-Q    & 1.16      & 3.17      & 4.35      & 4.74  \\
    log-Q       & N/A       & 1.17      & 1.16      & 4.15  \\
    tanh-Q      & N/A       & 1.28      & 1.20      & 1.13   \\
    \textbf{QGAN}  & \textbf{3.32} & \textbf{4.15} & \textbf{4.46} & \textbf{4.37}   \\
    \bottomrule 
    \end{tabular}
    \end{sc}
    \end{small}
%\vskip -0.1in
\end{table}

%To demonstrate the impact of the proposed QGAN, we take the aforementioned three quantization schemes as competitors, i.e. minmax quantization (minmax-Q), logarithmic minmax quantization (log-Q), and tanh quantization (tanh-Q). The results of quantized DCGAN from 1-bit to 4-bit representations on CIFAR-10 are shown in Table \ref{tab:best_score}, and the IS here are the obtained best score. To simplify the comparisions, we quantize the discriminator and the generator to same bits. The lost points in 1-bit cases of log-Q and tanh-Q are due to their degeneration. 

\begin{figure}
    \centering
    \subfigure[2-bit QGAN in D]{
        \includegraphics[width=0.45\linewidth]{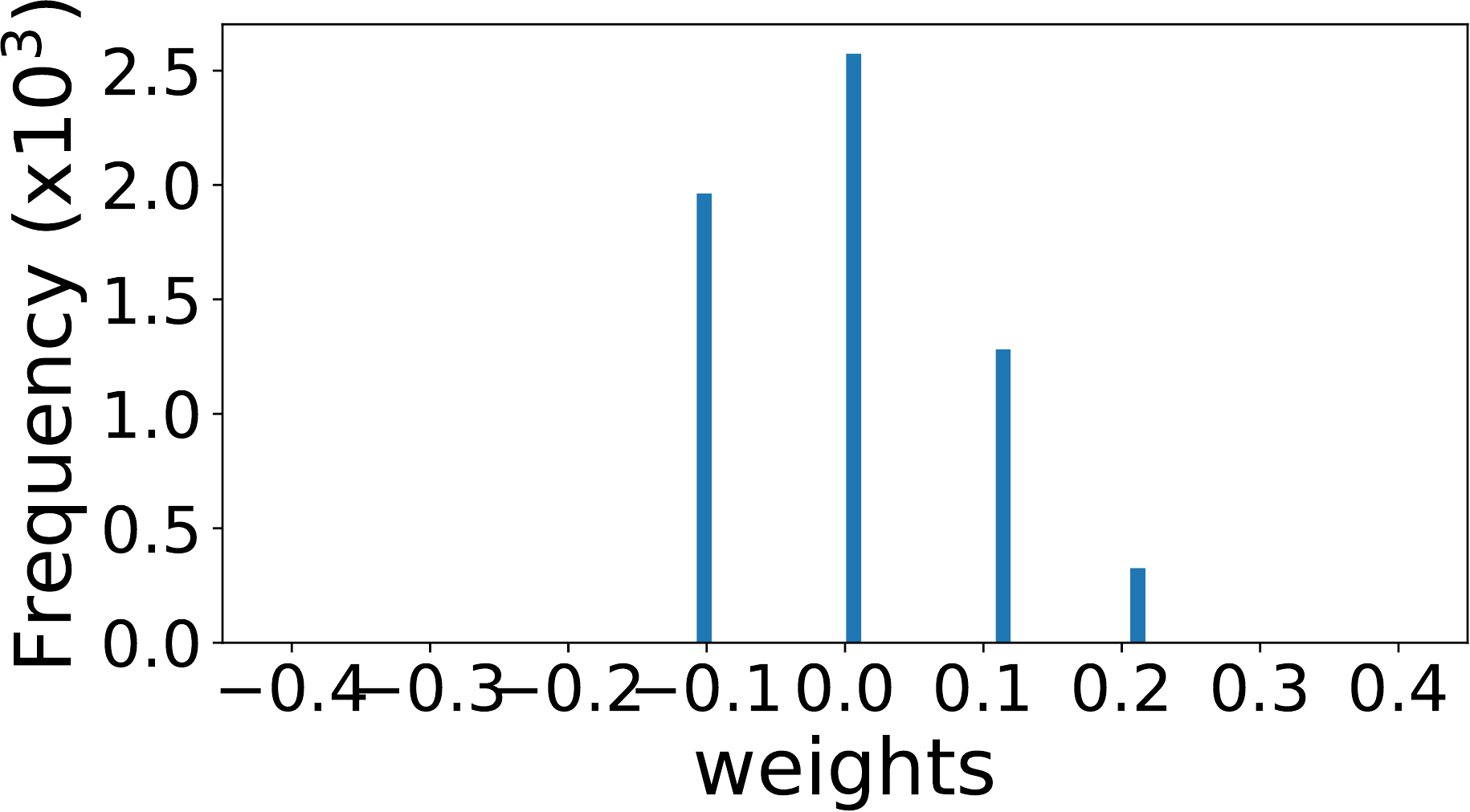}
    }
    \subfigure[2-bit QGAN in G]{
        \includegraphics[width=0.45\linewidth]{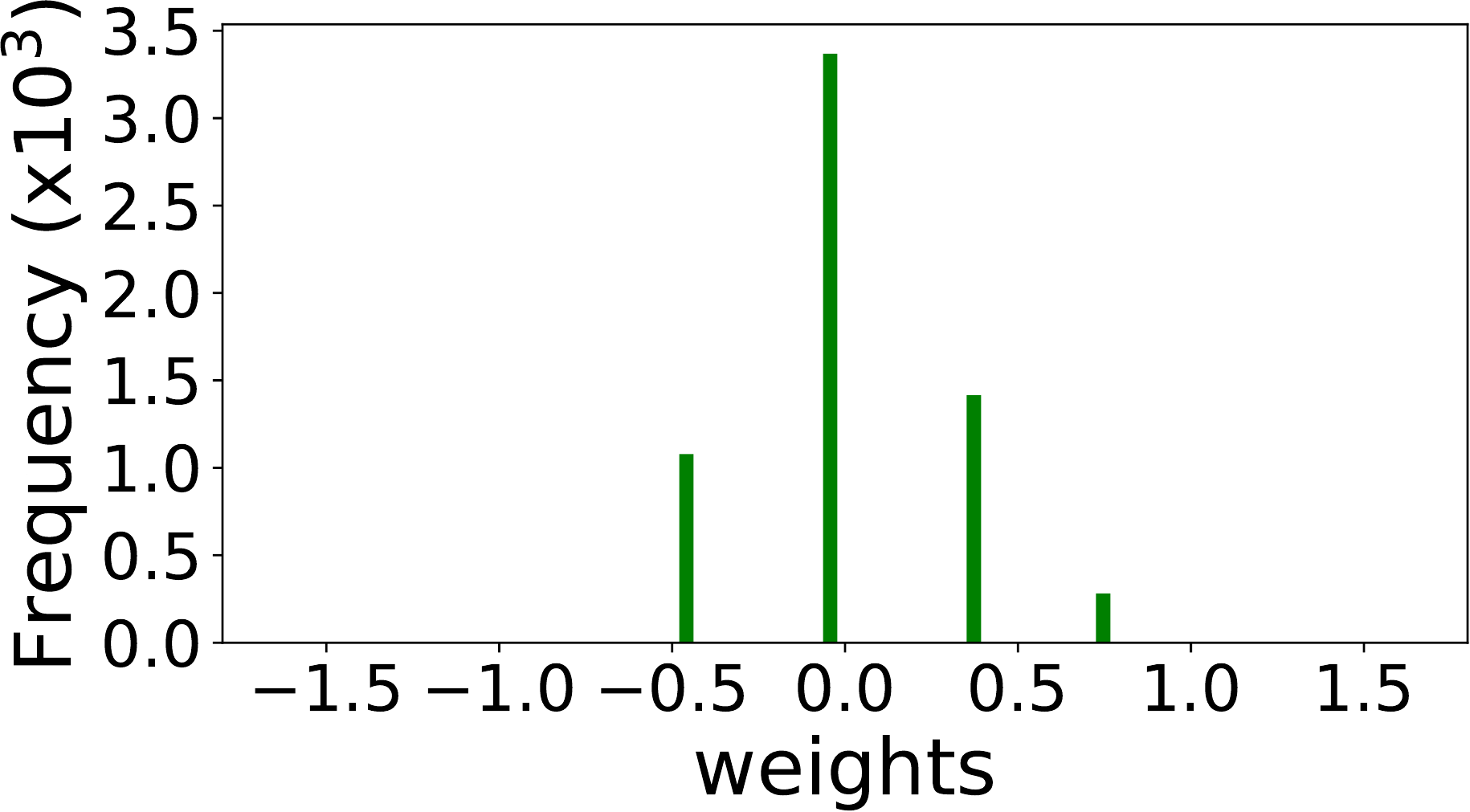}
    }
    \vskip -0.1in
    \caption{The distribution of weights in quantized DCGAN using 2-bit QGAN.}
    \vskip -0.1in
    \label{fig:distribution_em}
\end{figure}

Results in Table~\ref{tab:best_score} show that QGAN gets the best or comparable results in all cases.
We inspect the distribution of quantized states in QGAN, which is shown in Figure~\ref{fig:distribution_em}.
Compared to Figure~\ref{fig:distribution}, quantization based on the EM algorithm can overcome the problem of data underrepresentation, thus resulting in a better fit of quantized states to the distribution of original weights.
Besides, these results also show that QGAN can still work in the case using extreme low-bit data representations, specifically 1-bit where GAN models become binary neural networks.
Although there is still a quality gap between the 1-bit model quantized by QGAN and the baseline full-precision model, all other quantization methods either fail or generate noise in this extreme case.

%Compared to other quantization methods, QGAN gets the best results. The distributions of 2-bit DCGAN using QGAN are shown in Figure \ref{fig:distribution_em}. Comparing with Figure \ref{fig:distribution}, it can be seen that QGAN fits the original distribution best.
%The small gap between 4-bit minmax-Q and QGAN is within the allowed range because the inception score is not always an exact metric like accuracy. Figure ~\ref{fig:IS} displays the generated samples in different Inception Scores. It can be seen that the images in the case $IS=2$ have obvious artifacts, while images in cases $IS=4$ and $IS=5$ have less visible difference.

%The linear transform we used as the scaling function in QGAN gets rid of the limitation of the input minimum and maximum, thus it is better with fitting data distributed in various ranges in GAN models. The issues such as quantized states degradation or waste, which appear in other quantization schemes, are addresses to some extent. Moreover, the EM algorithm can find the proper scaling parameters, helping our quantization scheme achieving high utilization of quantized states.

%The results in Table \ref{tab:best_score} also show that QGAN can still work even in extreme low-bit case, i.e. 1-bit quantization. In thus a case, the GAN becomes the binary version. Although the results of 1-bit QGAN have a quality gap with the baseline IS score (full precision version) shown in Table \ref{table:2-bit}, all other quantization methods fail in this quantize case. 

\begin{figure*} [!t]
    \centering
    \subfigure[DCGAN baseline]{
        \label{subfig:dcgan_base}
        \includegraphics[width=0.21\linewidth]{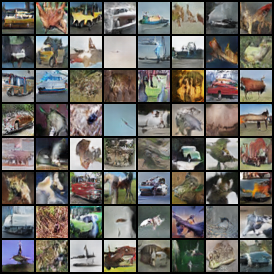}
    }
    \subfigure[LSGAN baseline]{
        \label{subfig:lsgan_base}
        \includegraphics[width=0.21\linewidth]{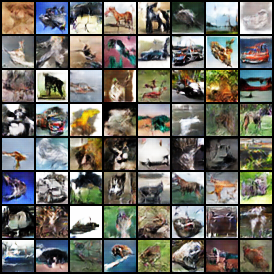}
    } 
    \subfigure[WGAN-GP baseline]{
        \label{subfig:wgan-gp_base}
        \includegraphics[width=0.21\linewidth]{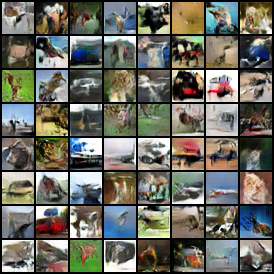}
    }     
    \subfigure[DCGAN baseline]{
        \label{subfig:dcgan_face}
        \includegraphics[width=0.21\linewidth]{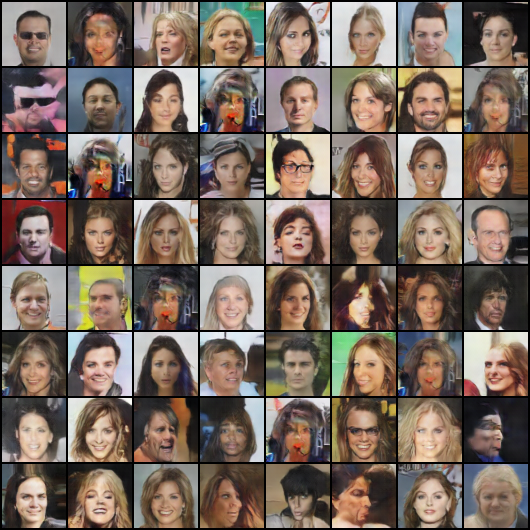}
    }  \\   
    \subfigure[DCGAN with 1D2G]{
        \label{subfig:dcgan_1d2g}
        \includegraphics[width=0.21\linewidth]{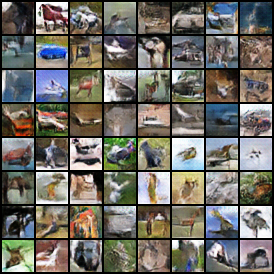}
    }
    \subfigure[LSGAN with 3D3G]{
        \label{subfig:lsgan_3d3g}
        \includegraphics[width=0.21\linewidth]{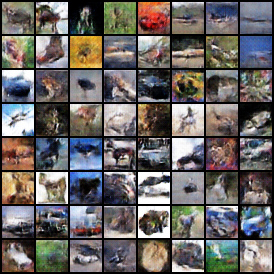}
    } 
    \subfigure[WGAN-GP with 4D4G]{
        \label{subfig:wgan-gp_4d4g}
        \includegraphics[width=0.21\linewidth]{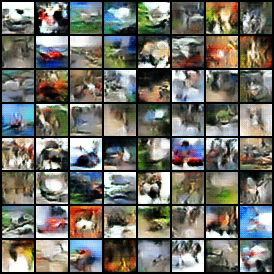}
    }
    \subfigure[DCGAN with 1D3G]{
        \label{subfig:dcgan_face_1d3g}
        \includegraphics[width=0.21\linewidth]{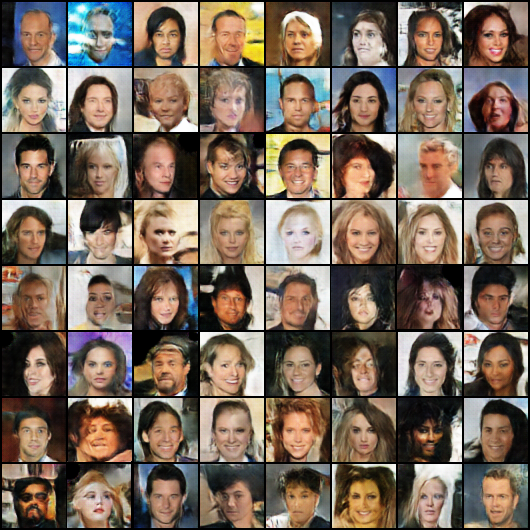}
    } 
    \vskip -0.1in
    \caption{The generated samples of various GAN models on CIFAR-10 dataset and DCGAN on CelebA dataset using QGAN. The kDjG denotes $k$-bit D and $j$-bit G.}
    \label{fig:generated_samples}
\end{figure*}

\begin{figure*}[!t]
    \centering
    \subfigure[Quantized D only]{
        \includegraphics[width=0.26\linewidth]{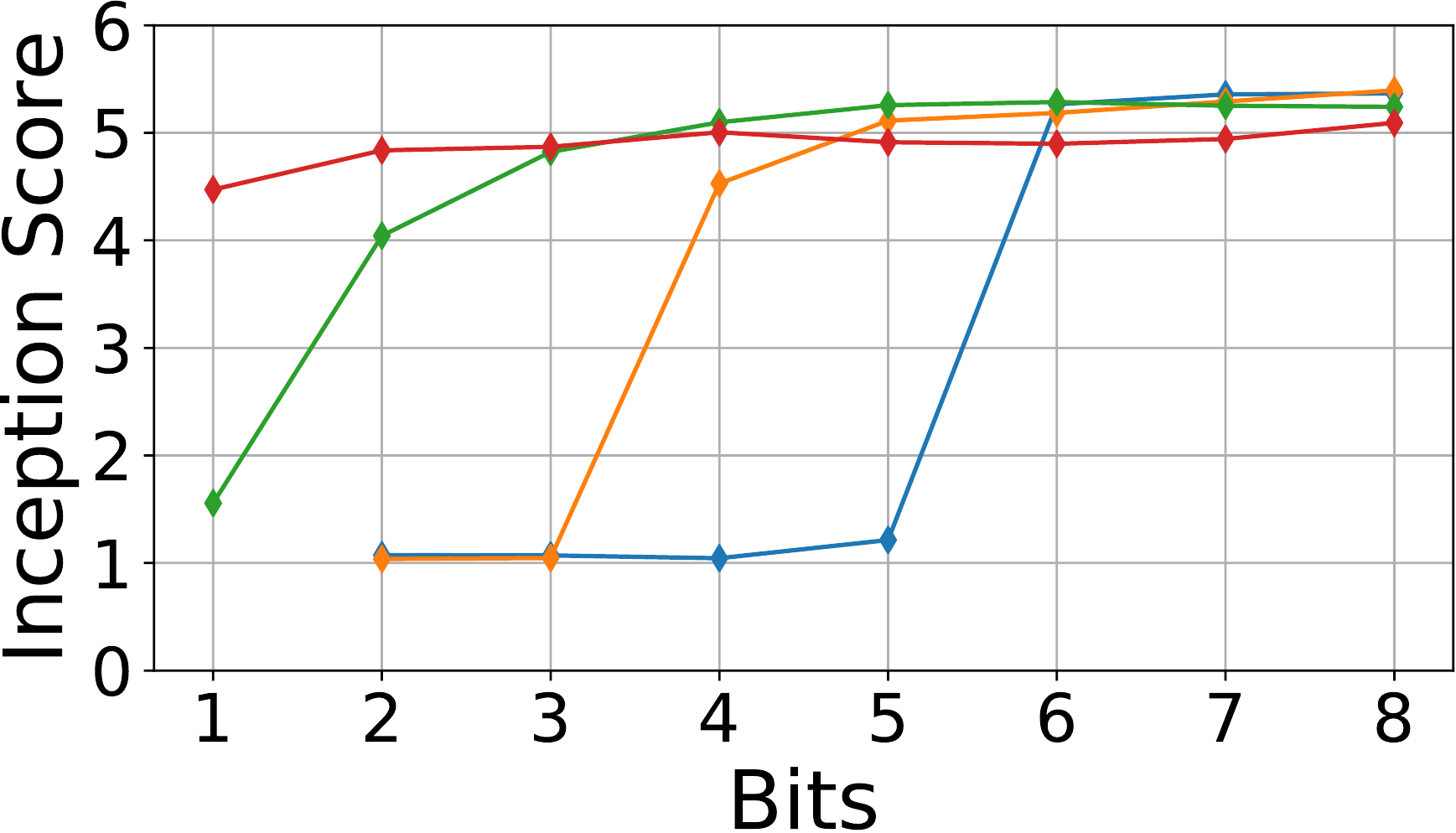}
        \label{fig:BS_D_DCGAN}
    } \quad
    \subfigure[Quantized both D and G]{
        \includegraphics[width=0.26\linewidth]{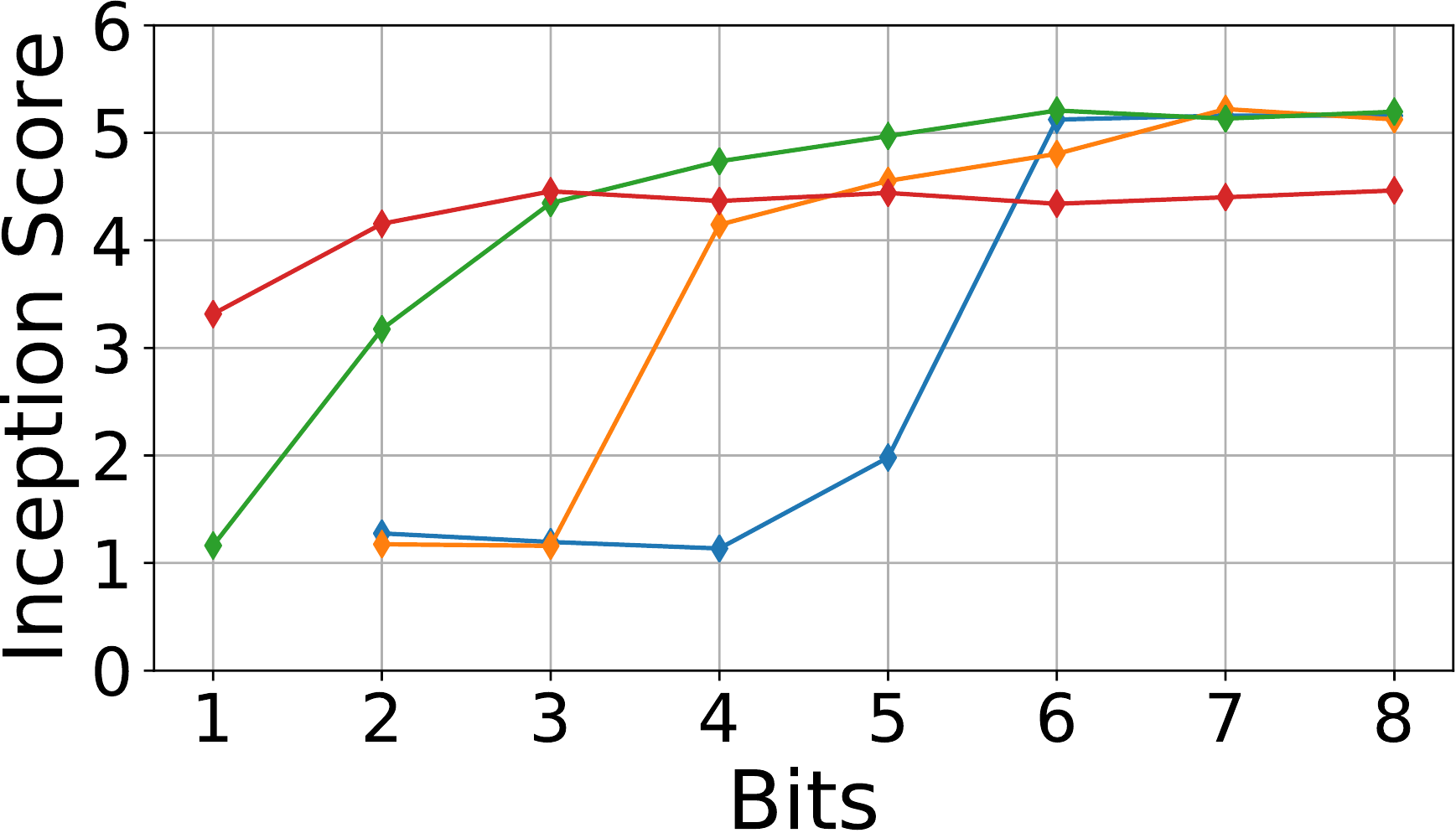}
        \label{fig:BS_D&G_DCGAN}
    } \quad
    \subfigure[Quantized G only]{
        \includegraphics[width=0.37\linewidth]{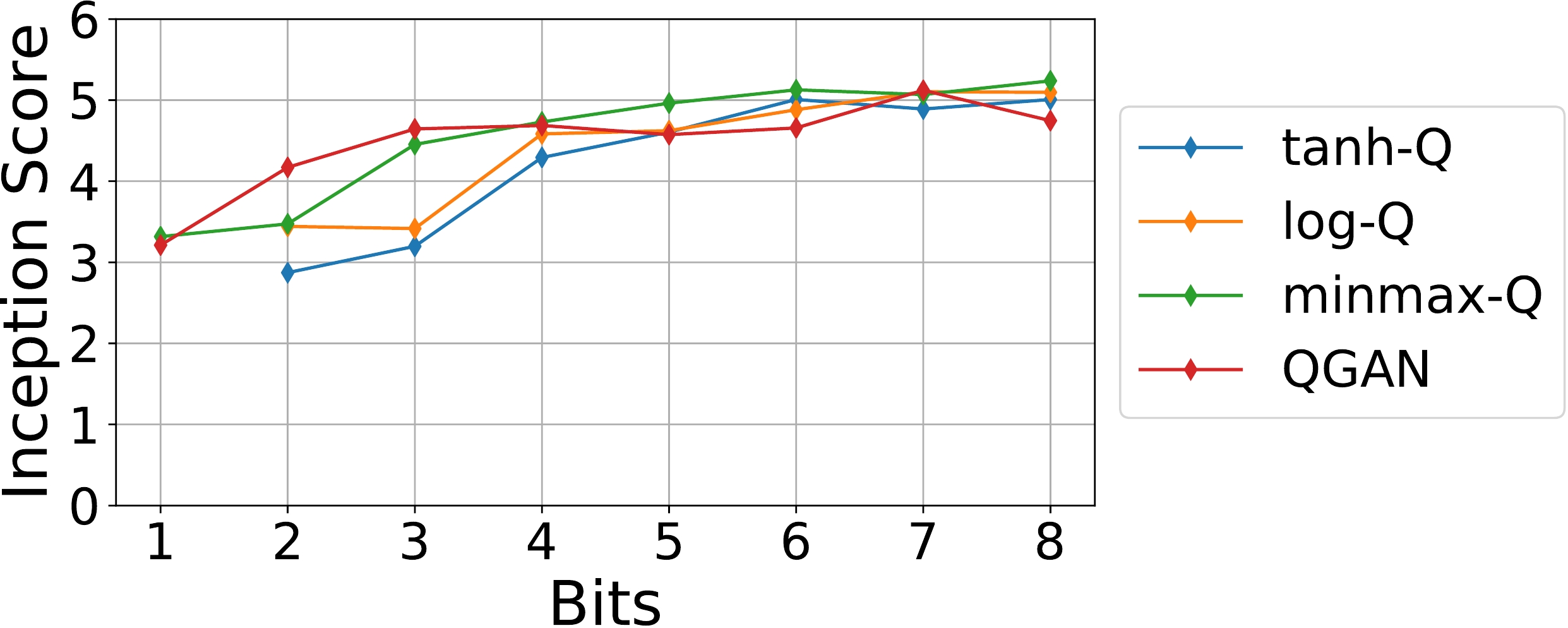}
        \label{fig:BS_G_DCGAN}
    }
    \vskip -0.1in
    \caption{The inception scores of generated samples of DCGANs with different bits using different quantization methods on CIFAR-10.}
    \label{fig:best_score}
    %\vskip -0.1in
\end{figure*}

\subsection{Multi-precision quantization} \label{sec:overall_results}

\begin{table}[!t]
    \centering
    \caption{The Inception Scores of different GAN models using multi-precision QGAN}
    \label{tab:overall}
    \vskip 0.15in
    \begin{small}
    \setlength\tabcolsep{3.5pt}
    \begin{sc}
    \begin{tabular}{lccccc}
    \toprule
    Model   & Dataset   & D-bit & G-bit & IS    & IS-32bits \\
    \midrule
    DCGAN   & CIFAR-10  & 1     & 2     & 4.33  & 5.30       \\
    WGAN-GP & CIFAR-10  & 4     & 4     & 3.17  & 4.31       \\
    LSGAN   & CIFAR-10  & 3     & 3     & 3.55  & 4.91       \\
    DCGAN   & CelebA    & 1     & 3     & 2.68  & 2.67       \\
    \bottomrule 
    \end{tabular}
    \end{sc}
    \end{small}
%\vskip -0.1in
\end{table}

To demonstrate the effectiveness of our multi-precision quantization, we apply it to three GAN models on two datasets.
Overall results are shown in Table~\ref{tab:overall}.
We also present the images generated by quantized models compared to images generated by their baseline in Figure~\ref{fig:generated_samples}.
Although IS reflects the quality of generated images to some extents, it is still hard to find a certain lowest bound of IS for the acceptable image quality.
Therefore, we classify generated images into three categories, acceptable, unacceptable, and unknown.
For experiments on CIFAR-10, we take images with IS larger than 4 as ones in acceptable quality, images with IS smaller than 3 as ones in unacceptable quality, and other cases as ones need a manual inspection for the image quality.
We show the generated images in Figure~\ref{fig:generated_samples}to demonstrate that our criteria are reasonable.
Besides, we would like to conclude that IS is related to the dataset used for the evaluation.
The IS of generated images from the baseline trained by CelebA is only 2.67 while the images shown in Figure~\ref{subfig:dcgan_face} and~\ref{subfig:dcgan_face_1d3g} are in reasonable quality.

Since our multi-precision quantization method is motivated by observations in Section~\ref{sec:sensitive}, we also examine the sensitivities of the discriminator and the generator for other quantization methods besides log-Q.
Figure~\ref{fig:best_score} presents the IS of quantized DCGAN on CIFAR-10 using different quantization methods under different number of bits.
Results shown in Figure~\ref{fig:best_score} confirm that our observations in Section~\ref{sec:sensitive} are applicable to other quantization methods.
Comparing Figure~\ref{fig:BS_D_DCGAN} and~\ref{fig:BS_D&G_DCGAN}, the quantized $D$ and quantized both $D$ and $G$ converge in cases with the same number of bits, i.e. 2-bit in QGAN and 4-bit in log-Q.
Comparing Figure~\ref{fig:BS_D&G_DCGAN} and~\ref{fig:BS_G_DCGAN}, once quantizing only $D$ can converge, the variation of the numebr of bits in $G$ has little impact on the whole GAN model.
These results validate the generality of our observations which also indicate the effectiveness of our multi-precision quantization process on other quantization methods.

\section{Conclusion}

%We proposed QGAN, which is the first to quantize generative adversarial networks (GANs) into low-bit representation with a resonable quality of generated samples. 
In this paper, we study the problem of quantizing generative adversarial networks (GANs).
We first conduct an extensive study on the effectiveness of typical quantization methods which are widely used in CNNs or RNNs. 
Our observation reveals that the underrepresentation of original values in quantized states leads to the failure of these methods in quantizing GAN. 
The observation motivates us to propose QGAN, which operates with a linear scaling function based on EM algorithm and achieves high utlilization of quantized states
%to overcome the issue of underrepresentation.
Besides, we observe from the case study that the discriminator is more sensitive than the generator to the number of quantized bits. 
To leverage this observation, we introduce a multi-precision quantization approach to find the lowest number of bits for quantizing GAN models to satisfy the quality requirement for generated samples.
Our experiments on various GANs and different datasets show that QGAN can generate samples in a comparable quality in cases using even only 1-bit or 2-bit.


\bibliography{ref}
\bibliographystyle{icml2019}

\end{document}